\documentclass[11pt,a4paper]{article}
\usepackage[hyperref]{emnlp-ijcnlp-2019}
\usepackage{times}
\usepackage{latexsym}
\usepackage{multirow}
\usepackage{graphicx}
\usepackage{tabularx}
\usepackage{url}
\usepackage{enumitem}
\usepackage{natbib}
\usepackage{subfig}
\usepackage{hyperref}
\usepackage{amsmath}
\usepackage{algorithm}
\usepackage[noend]{algpseudocode}

\aclfinalcopy 

\title{CogniVal: A Framework for Cognitive Word Embedding Evaluation}

\author{Nora Hollenstein\textsuperscript{1}, Antonio de la Torre\textsuperscript{1}, Nicolas Langer\textsuperscript{2}, Ce Zhang\textsuperscript{1} \\
  \textsuperscript{1} Department of Computer Science, ETH Zurich\\
  {\tt \{noraho,antonide,ce.zhang\}@ethz.ch} \\
  \textsuperscript{2} Department of Psychology, University of Zurich \\
  {\tt n.langer@psychologie.uzh.ch} \\}
\date{}

\begin{document}
\maketitle
\begin{abstract}
An interesting method of evaluating word representations is by how much they reflect the semantic representations in the human brain. However,
most, if not all, previous works only focus on small datasets and a single modality. In this paper,
we present the first multi-modal framework for evaluating English word representations based on cognitive lexical semantics. Six types of word embeddings are evaluated by fitting them to 15 datasets of eye-tracking, EEG and fMRI signals recorded during language processing. To achieve a global score over all evaluation hypotheses, we apply statistical significance testing accounting for the multiple comparisons problem. This framework is easily extensible and available to include other intrinsic and extrinsic evaluation methods. We find strong correlations in the results between cognitive datasets, across recording modalities and to their performance on extrinsic NLP tasks.
\end{abstract}

\section{Introduction}

Word embeddings are the corner stones of state-of-the-art NLP models. Distributional representations which interpret words, phrases, and sentences as high-dimensional vectors in semantic space have become increasingly popular. These vectors are obtained by training language models on large corpora to encode contextual information. Each vector represents the meaning of a word. 

\begin{figure}[h]
    \centering
    \includegraphics[width=0.4\textwidth]{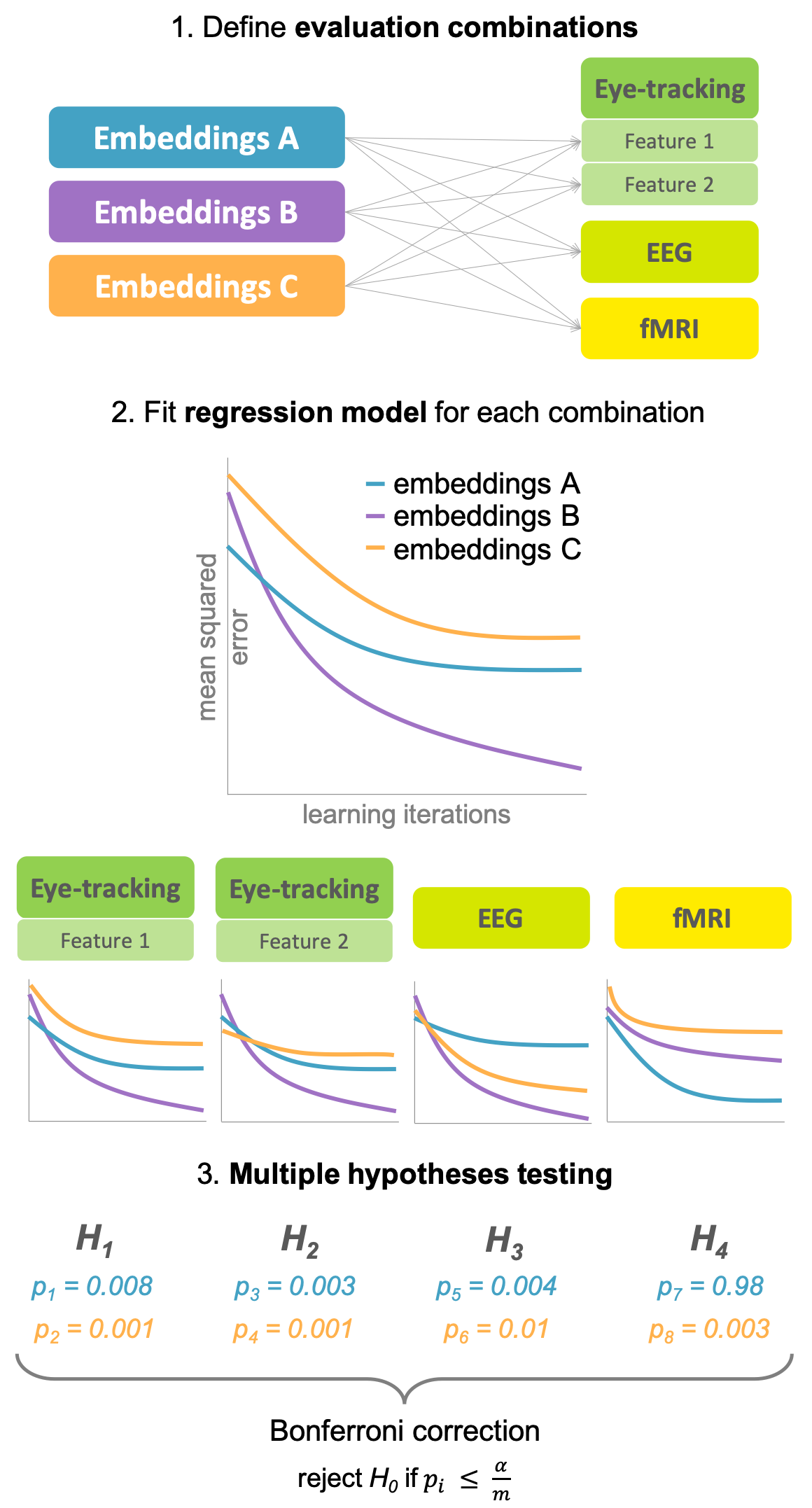}
    \caption{Overview of the cognitive word embedding evaluation process.}
    \label{fig:sysoverview}
\end{figure}

Evaluating and comparing the quality of different word embeddings is a well-known, largely open challenge. Currently, word embeddings are evaluated with extrinsic or intrinsic methods. Extrinsic evaluation is the process of assessing the quality of the embeddings based on their performance on downstream NLP tasks, such as question answering or entity recognition. However, embeddings can be trained and fine-tuned for specific tasks, but this does not mean that they accurately reflect the meaning of words.

One the other hand, intrinsic evaluation methods, such as word similarity and word analogy tasks, merely test single linguistic aspects. These tasks are based on conscious human judgements. Conscious judgements can be biased by subjective factors and the tasks themselves might also be biased \cite{nissim2019fair}.
Additionally, the correlation between intrinsic and extrinsic metrics is not very clear, as intrinsic evaluation results fail to predict extrinsic performance \cite{chiu2016intrinsic,gladkova2016intrinsic}. Finally, both intrinsic and extrinsic evaluation types often lack statistical significance testing and do not provide a global quality score.

In this paper, we focus on the {\em intrinsic subconscious evaluation method} \cite{bakarov2018survey}, {\em which evaluates English word embeddings against the lexical representations of words in the human brain, recorded when passively understanding language}. Cognitive lexical semantics proposes that words are defined by how they are organized in the brain \cite{miller1992wordnet}. As a result, brain activity data recorded from humans processing language is arguably the most accurate mental lexical representation available \cite{sogaard2016evaluating}. Recordings of brain activity play a central role in furthering our understanding of how human language works. To accurately encode the semantics of words, we believe that embeddings 
should reflect this mental lexical representation.

Evaluating word embeddings with cognitive language processing data has been proposed previously. However, the available datasets are not large enough for powerful machine learning models, the recording technologies produce noisy data, and most importantly, only few datasets are publicly available. Furthermore, since brain activity and eye-tracking data contain very noisy signals, correlating distances between representations does not provide sufficient statistical power to compare embedding types \cite{frank2017word}. For this reason we evaluate the embeddings by exploring how well they can predict human processing data. We build on \citet{sogaard2016evaluating}'s theory of evaluating embeddings with this task-independent approach based on cognitive lexical semantics and examine its effectiveness. The design of our framework
follows three principles:
\begin{enumerate}
  \setlength{\itemsep}{1pt}
  \setlength{\parskip}{0pt}
  \setlength{\parsep}{0pt}
    \item \textbf{Multi-modality}: Evaluate against various modalities of recording human signals to counteract the noisiness of the data.
    \item \textbf{Diversity} within modalities: Evaluate against different datasets within one modality to make sure the number of samples is as large as possible.
    \item \textbf{Correlation} of results should be evident across modalities and even between datasets of the same modality.
\end{enumerate}

\paragraph{Contributions} We present CogniVal, the first framework of cognitive word embedding evaluation to follow these principles and analyze the findings. We evaluate different embedding types against a combination of 15 cognitive data sources, acquired via three modalities: eye-tracking, electroencephalography (EEG) and functional magnetic resonance imaging (fMRI). The word representations are evaluated by assessing their ability of predicting cognitive language processing data. After fitting a neural regression model for each combination, we apply multiple hypotheses testing to measure the statistical significance of the results, taking into account multiple comparisons (see Figure \ref{fig:sysoverview}). This contributes to the consistency of the results and to attain a global score of embedding quality. Our main findings when evaluating six state-of-the-art word embeddings with CogniVal show that the majority of embedding types significantly outperform a baseline of random embeddings when predicting a wide range of cognitive features. Additionally, the results show consistent correlations between between datasets of the same modality and across different modalities, validating the intuition of our approach. Finally, we present an exploratory but promising correlation analysis between the scores obtained using our intrinsic evaluation methods
and the performance on extrinsic NLP tasks.

The code of this evaluation framework is openly available\footnote{\url{https://github.com/DS3Lab/cognival}}. It can be used as is, or in combination with other intrinsic as well as extrinsic evaluation methods for word representations.

\section{Related Work}



\citet{mitchell2008predicting} pioneered the use of word embeddings to predict patterns of neural activation when subjects are exposed to isolated word stimuli. More recently, this dataset and other fMRI resources have been used to evaluate learned word representations. 

For instance, \citet{abnar2018experiential} and \citet{rodrigues2018predicting} evaluate different embeddings by predicting the neuronal activity from the 60 nouns presented by \citet{mitchell2008predicting}. \citet{sogaard2016evaluating} shows preliminary results in evaluating embeddings against continuous text stimuli in eye-tracking and fMRI data. Moreover, \citet{beinborn2019robust} recently presented an extensive set of language--brain encoding experiments. Specifically, they evaluated the ability of an ELMo language model to predict brain responses of multiple fMRI datasets. 

EEG data has been used for similar purposes. \citet{schwartz2019understanding} and \citet{ettinger2016modeling} show that components of event-related potentials can successfully be predicted with neural network models and word embeddings.

However, these approaches mostly focus on one modality of brain activity data from small individual cognitive datasets. The lack of data sources has been one reason why this type of evaluation has not been too popular until now \cite{bakarov2018can}. Hence, in this work we collected a wide range of cognitive data sources ranging from eye-tracking to EEG and fMRI to ensure coverage of different features, and consequently of the cognitive processing taking place in the human brain during reading.

\paragraph{Evidence from cognitive neuroscience}

\citet{murphy2018decoding} review computational approaches to the study of language with neuroimaging data and show how different type of words activate neurons in different brain regions. Similarly, mapping fMRI data from subjects listening to stories to the activated brain regions, revealed semantic maps of how words are distributed across the human cerebral cortex \cite{huth2016natural}.

Furthermore, word predictability and semantic similarity show distinct patterns of brain activity during language comprehension: semantic distances can have neurally distinguishable effects during language comprehension \cite{frank2017wordpred}. These findings support the theory that brain activity data does reflect lexical semantics and is thus an appropriate foundation for evaluating the quality of word embeddings.

\section{Word embeddings}

Pre-trained word vectors are an essential component in state-of-the-art NLP systems. 
We chose six commonly used pre-trained embeddings to evaluate against the cognitive data sources. See Table \ref{embed} for an overview of the dimensions of each embedding type. We evaluate the following types of word embeddings:

\begin{table}[t]
\centering
\begin{tabular}{lcc}
\hline
\textbf{embeddings} & \textbf{dim.} & \textbf{hidden layer units} \\\hline\hline
Glove & 50  & [30, 26, 20, 5]\\
Glove & 100  & [50, 30]\\
Glove & 200  & [100, 50] \\
Glove & 300  & [150, 50]\\\hline
Word2vec & 300  & [150, 50] \\\hline
WordNet2vec & 850  &  [400, 200] \\\hline
FastText & 300  & [150, 50]  \\\hline
ELMo & 1024  & [600, 200]  \\\hline
BERT & 768  & [400, 200] \\
BERT & 1024  & [600, 200]\\\hline
\end{tabular}
\caption{Overview of word embeddings evaluated with CogniVal. The last column shows the search space of the grid search for the number of units in the hidden layer.}
\label{embed}
\end{table}

\begin{itemize}

    \item \textbf{Glove}: \citet{pennington2014glove} provide embeddings of different dimensions trained on aggregated global word-word co-occurrence statistics over a corpus of 6 billion words.
    \newpage
    \item \textbf{Word2vec}: Non-contextual embeddings trained on 100 billion words from a Google News dataset \cite{mikolov2013distributed}.
    \item \textbf{WordNet2Vec} \cite{saedi2018wordnet} These embeddings represent the conversion from semantic networks into semantic spaces. Trained on WordNet, a lexical ontology for English that comprises over 155,000 lemmas (but trained only on 60,000 words).
    \item \textbf{FastText} pre-trained embeddings use character n-grams to compose the vector of the full words \cite{mikolov2018advances}. We evaluate the embeddings with and without subword information trained on 16 billion tokens of Wikipedia sentences as well as the ones trained on 600 billion tokens of Common Crawl.
    \item \textbf{ELMo} models both complex characteristics of word use (i.e. syntax and semantics), and how these uses vary across linguistic contexts \cite{peters2018deep}. These word vectors are learned functions of the internal states of a deep bidirectional language model, which is pre-trained on a large text corpus. We take the first of the three output layers, containing the context insensitive word representations.
    \item \textbf{BERT} embeddings are contextual, bidirectional word representations, based on the idea that fine-tuning a pre-trained language model can help the model achieve better results in the downstream tasks \cite{devlin2019bert}. We take the hidden states of the second to last of 12 output layers as the representation for each token.
    
\end{itemize}

\section{Cognitive data}\label{data}

\begin{table*}[t]
\centering
\begin{tabular}{|l|lccccc|}
\hline
 & \textbf{Data source} & \textbf{stimulus} & \textbf{subj.}  & \textbf{tokens} & \textbf{types} & \textbf{coverage} \\\hline
 &
 GECO \cite{cop2017presenting} & text & 14 & 68606 & 5383 & 95\% \\
 \parbox[t]{2mm}{\multirow{3}{*}{\rotatebox[origin=c]{90}{\textsc{Eye-Tracking}}}} & \textsc{Dundee} \cite{kennedy2003dundee} & text  & 10 & 58598 & 9131 & 94\% \\
 & \textsc{CFILT-Sarcasm} \cite{mishra2016predicting} & text  & 5 & 23466 & 4237 & 85\% \\
 & \textsc{ZuCo} \cite{hollenstein2018zuco} & text & 12 &  13717 & 4384 & 90\% \\
   & \textsc{CFILT-Scanpath} \cite{mishra2017scanpath} & text  & 5 & 3677 & 1314 & 89\% \\
 & \textsc{Provo} \cite{luke2017provo} & text & 84 & 2743 & 1192 & 95\% \\
 & UCL \cite{frank2013reading} & text & 43 & 1886 & 711 & 98\% \\
 & \textsc{all eye-tracking} (aggregated)  & text & - & 26353 & 16419 & 88\% \\\hline\hline
\parbox[t]{2mm}{\multirow{3}{*}{\rotatebox[origin=b]{90}{\textsc{EEG}}}} & ZuCo \cite{hollenstein2018zuco} & text & 12 & 
 13717 & 4384 & 90\% \\
 & \textsc{Natural Speech} \cite{broderick2018electrophysiological} & speech   & 19 & 12000 & 1625 & 98\% \\
 & UCL \cite{frank2015erp} & text & 24 & 1931 & 711 & 98\% \\
 & \textsc{N400} \cite{broderick2018electrophysiological} & text   & 9 & 150 & 140 & 100\% \\
\hline\hline
\parbox[t]{2mm}{\multirow{3}{*}{\rotatebox[origin=b]{90}{fMRI}}} & \textsc{Harry Potter} \cite{wehbe2014aligning} & text & 8 & 5169 & 1295 & 92\% \\
 & \textsc{Alice} \cite{brennan2016abstract} & speech  & 27 & 2066 & 588 & 99\% \\
  & \textsc{Pereira} \cite{pereira2018toward} & text/image   & 15 & 180 & 180 & 99\% \\
 & \textsc{Nouns} \cite{mitchell2008predicting} & image   & 9 & 60 & 60 & 100\%\\
\hline
\end{tabular}
\caption{Cognitive data sources used in this work. Coverage is the percentage of vocabulary in data source occurs in British National Corpus list of most frequent English words\footnotemark .}
\label{cognitive-data}
\end{table*}

In this paper, we consider three modalities of recording cognitive language processing signals: eye-tracking, electroencephalography (EEG), and functional magnetic resonance imaging (fMRI). All three methods are complementary in terms of temporal and spatial resolution as well as the directness in the measurement of neural activity \cite{mulert2013simultaneous}. 
For the word embedding evaluation we selected a wide range of datasets from these three modalities to ensure a more diverse and accurate representation of the brain activity during language processing.

Table \ref{cognitive-data} shows an overview of the cognitive data sources used, which are described in more detail below. Since the processing in the brain differs depending on whether the information is accessed via the visual or auditory system \cite{price2012review}, we include data of different stimuli, e.g. participants reading sentences or listening to audio-books. Moreover, our collection of cognitive data sources contains datasets of both isolated (single words) and continuous (words in context, i.e. sentences or stories) stimuli. All datasets include English language stimuli and the participants were native speakers or highly proficient.

\paragraph{Eye-tracking}
Eye-tracking is an indirect measure of cognitive activity. Gaze patterns are highly correlated with the cognitive load associated with different stages of human text processing \cite{rayner1998eye}. For instance, fixation duration is higher for long, infrequent and unfamiliar words \cite{just1980theory}.

All eye-tracking datasets used in this work were recorded from natural, self-paced reading. Each dataset provides different eye-tracking features. The most common features, available in all 7 datasets are: first fixation duration, first pass duration, mean fixation duration, total fixation duration and number of fixations. For a complete list and description of the eye-tracking features available in each corpus see Appendix \ref{app-gaze}.

Gaze vectors consist of specific features, which are extracted based on the reading times, fixations and regressions on each word. Feature values are aggregated on word type level and scaled between 0 and 1. The feature values were averaged over all subjects within a dataset. This preprocessing step is done separately for each data source before combining them. \citet{hollenstein2019entity} show that combining gaze data from different sources can be helpful for NLP applications, even when they are recorded with different devices and filtering,

By using as many features as available from each dataset, including features characterizing basic, early and late word processing aspects, the goal is to cover the whole language understanding process on word level.

\paragraph{EEG}
Electroencephalography records electrical activity from the brain. It measures voltage fluctuations through the scalp with high temporal resolution.
\citet{hauk2004effects} present evidence for the modulation of early electrophysiological brain responses by word frequency. This is evidence that lexical access from written word stimuli is an early process that follows stimulus presentation by less than 200 ms.

The EEG datasets used in this work were either recorded from reading sentences or listening to natural speech. Word-level brain activity could be extracted by mapping to eye-tracking cues (\textsc{ZuCo}), by mapping to auditory triggers (\textsc{Natural Speech}), by recording only the last word in each sentence (N400), or through serial presentation of the words (UCL). Standard preprocessing steps for EEG data, including band-pass filtering and artifact removal, are performed in the same manner for all four data sources. See Appendix \ref{app-eeg} for details on EEG preprocessing.

The EEG data is aggregated over all available subjects and over all occurrences of a token. This yields an \textit{n}-dimensional vector, where \textit{n} is the number of electrodes, ranging from 32 to 130, depending on the EEG device used to record the data.

EEG data can be aggregated over all subjects within one dataset, because the number and locations of electrodes are identical. However, due to the differences in the number of electrodes between datasets, we cannot aggregate over all EEG datasets.

\paragraph{fMRI}

Functional magnetic resonance imaging is a technique for measuring and mapping brain activity by detecting changes associated with blood flow. fMRI has a temporal resolution of two seconds, which means that with continuous stimuli such as natural reading or story listening, one scan covers multiple words. We use datasets of isolated stimuli (e.g the \textsc{Nouns} dataset) as well as continuous stimuli (e.g. \textsc{Harry Potter}). While it is easier to extract word-level signals from isolated stimuli, continuous stimuli allow extracting signals in context over a wider vocabulary.

\footnotetext{\url{https://www.kilgarriff.co.uk/bnc-readme.html}}

Where multiple trials were available, the brain activation for each word is calculated by taking the mean over the scans.
Moreover, if the stimulus is continuous (\textsc{Harry Potter} and \textsc{Alice} datasets), the data is aligned with an offset of four seconds to account for hemodynamic delay\footnote{The fMRI signal measures a brain response to a stimulus with a delay of a few seconds, and it decays slowly over a duration of several seconds \cite{miezin2000characterizing}. For continuous stimuli, this means that the response to previous stimuli will have an influence on the current signal. Thus, context of the previous words is taken into account}. 

fMRI data contains representations of neural activity of millimeter-sized cubes called voxels. Standard fMRI preprocessing methods such as motion correction, slice timing correction and co-registration had already been applied before. To select the voxels to be predicted we use the pipeline provided by \citet{beinborn2019robust}. This pipeline consists of extracting corresponding scan(s) for each word, and randomly selecting 100, 500 and 1000 voxels (for the \textsc{Harry Potter}, \textsc{Pereira} and \textsc{Nouns} datasets). 
The published version of the \textsc{Alice} dataset provided the preprocessed signal averaged for six regions of interest, hence for this particular dataset we predict the activation for these regions only. Appendix \ref{app-fmri} contains the details of the preprocessing steps.
Finally, the fMRI data is converted to \textit{n}-dimensional vectors, where \textit{n} is the number of randomly selected voxels (100, 500 or 1000) or regions (6).

\section{Embedding evaluation method}

In order to evaluate the word embeddings against human lexical representations, we fit the embeddings to a wide range cognitive features, i.e. eye-tracking features and activation levels of EEG and fMRI. This section describes how these models were trained and evaluated. After evaluating each combination separately, we test for statistical significance taking into account the multiple comparisons problem. See Figure \ref{fig:sysoverview} for an overview of the evaluation process.

\begin{figure}[h]
    \centering
    \includegraphics[width=0.4\textwidth]{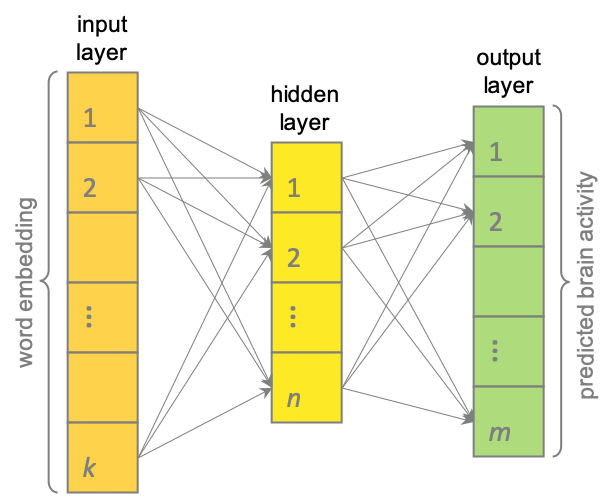}
    \caption{Neural architecture of regression models.}
    \label{fig:model}
\end{figure}

\subsection{Models}

We fit neural regression models to map word embeddings to cognitive data sources. Predicting multiple features from different sources and modalities allows us to evaluate different aspects of capturing the semantics of a word. Hence, separate models are trained for all combinations. For instance, fitting FastText embeddings to EEG vectors from \textsc{ZuCo}, or fitting ELMo embeddings to first fixation durations of the \textsc{Dundee} corpus.

\begin{figure*}[t]
    \centering
    \includegraphics[width=1\textwidth]{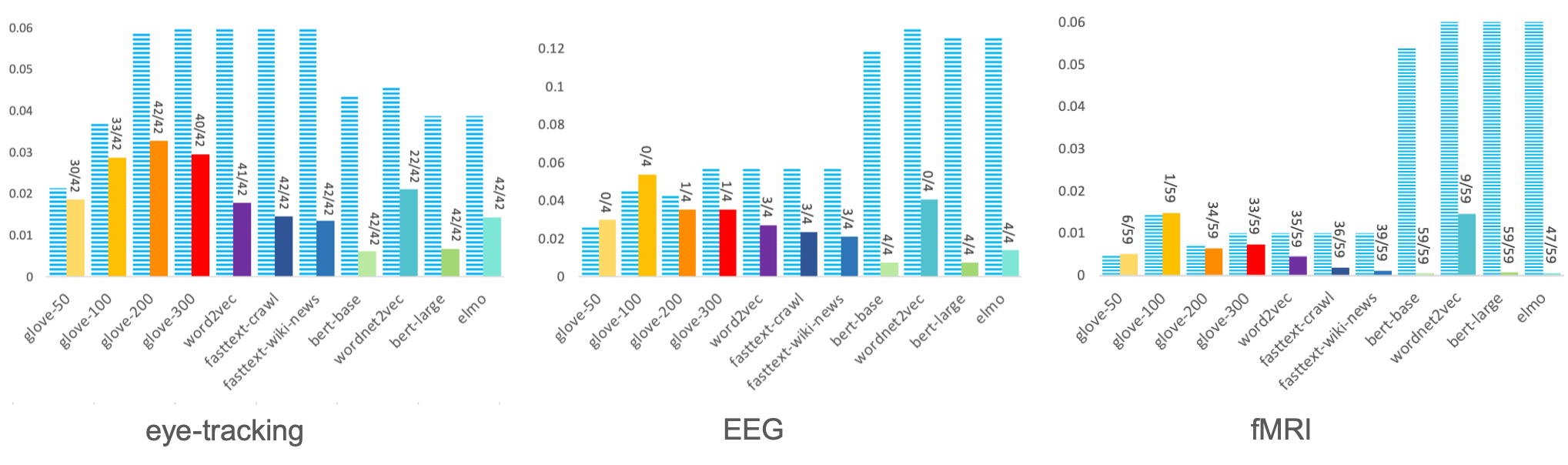}
    \caption{Results for each modality: Aggregated results for all embeddings predicting cognitive features for all datasets of a modality (sorted by dimension of embeddings in increasing order from left to right). The striped blued bars represent random baseline. The labels on the embedding bars show the ration of significant results under the Bonferroni correction to the total number of hypotheses.}
    \label{fig:results-ind}
\end{figure*}

For the regression models, we train neural networks with \textit{k} input dimensions, one dense hidden layer of \textit{n} nodes using ReLU activation and an output layer of \textit{m} nodes using linear activation. The model is a multiple regression with layers of dimension \textit{k-n-m}, where \textit{k} is the number of dimensions of the word embeddings and \textit{m} changes depending on the cognitive data source to be predicted. For predicting single eye-tracking features \textit{m} equals 1, whereas for predicting EEG of fMRI vectors \textit{m} is the dimension of the cognitive data vector, or more specifically, the number of electrodes in the EEG data or the number of voxels in the fMRI data. Figure \ref{fig:model} shows this neural architecture. The loss function optimizes the mean squared error (MSE) and uses an Adam optimizer with a learning rate of 0.001.

5-fold cross validation is performed for each model (80\% training data and 20\% test data).
The optimal number of nodes \textit{n} in the hidden layer is selected individually for each combination of cognitive data source and embedding type. To this end, a grid search is performed before training, which is evaluated on a validation set consisting of 20\% of the training data with 3-fold cross validation (see Table \ref{embed} for details on the search space). The best model is then saved and used to predict the cognitive feature for each word in the test set. Finally, the results are measured with the mean squared error, averaged over all predicted words. 

\begin{table}[t]
\centering
\begin{tabular}{|l|lll|}
\hline
 & \multicolumn{3}{c|}{\textbf{voxels}} \\
\textbf{embeddings} & 100 & 500 & 1000 \\\hline\hline
glove-300 & 0.119 & 0.081 & 0.078 \\
word2vec & 0.103 & 0.075 & 0.075 \\
fasttext-crawl-sub & 0.092 & 0.070 & 0.069 \\
bert-base & 0.020 & 0.017 & 0.016 \\
wordnet2vec & 0.105 & 0.077 & 0.076 \\
elmo & 0.067 & 0.051 & 0.050\\\hline
\end{tabular}
\caption{Effect of predicting different numbers of randomly selected voxels.}
\label{tab:voxels}
\end{table}

CogniVal allows for evaluation against another word embedding type as well as evaluation against a random baseline.
To generate a fair baseline we create random vectors for each word of \textit{n} dimensions, corresponding to the same number of dimensions of the embeddings to be evaluated. 

\begin{table}[t]
\centering
\begin{tabular}{|l|lll|}
\hline
\textbf{embeddings} & \textbf{nFix} & \textbf{TRT} & \textbf{FFD}\\\hline\hline
glove-300 & 0.010 & 0.017 & 0.027 \\
word2vec & 0.009  & 0.010 & 0.016\\
fasttext-crawl-sub & 0.008  & 0.007 & 0.012\\
bert-base & 0.005  & 0.003 & 0.004\\
wordnet2vec & 0.010  & 0.010 & 0.019\\
elmo & 0.008  & 0.009 & 0.011\\\hline
\textbf{average} & \textbf{0.008}  & \textbf{0.009}& \textbf{0.015}\\\hline
\end{tabular}
\caption{Comparison of word embeddings predicting single eye-tracking features: number of fixations (nFix), first fixation duration (FFD) and total reading time of a word (TRT).}
\label{tab:gaze}
\end{table}

\subsection{Multiple hypotheses testing}

With the purpose of achieving consistency and going towards a global quality metric that can be combined with other evaluation methods, we perform statistical significance testing on each hypothesis. A hypothesis consists of comparing the combination of an embedding type and a cognitive data source to the random baseline.

\begin{figure*}[t]
    \centering
    \subfloat[][]{\includegraphics[width=0.28\textwidth]{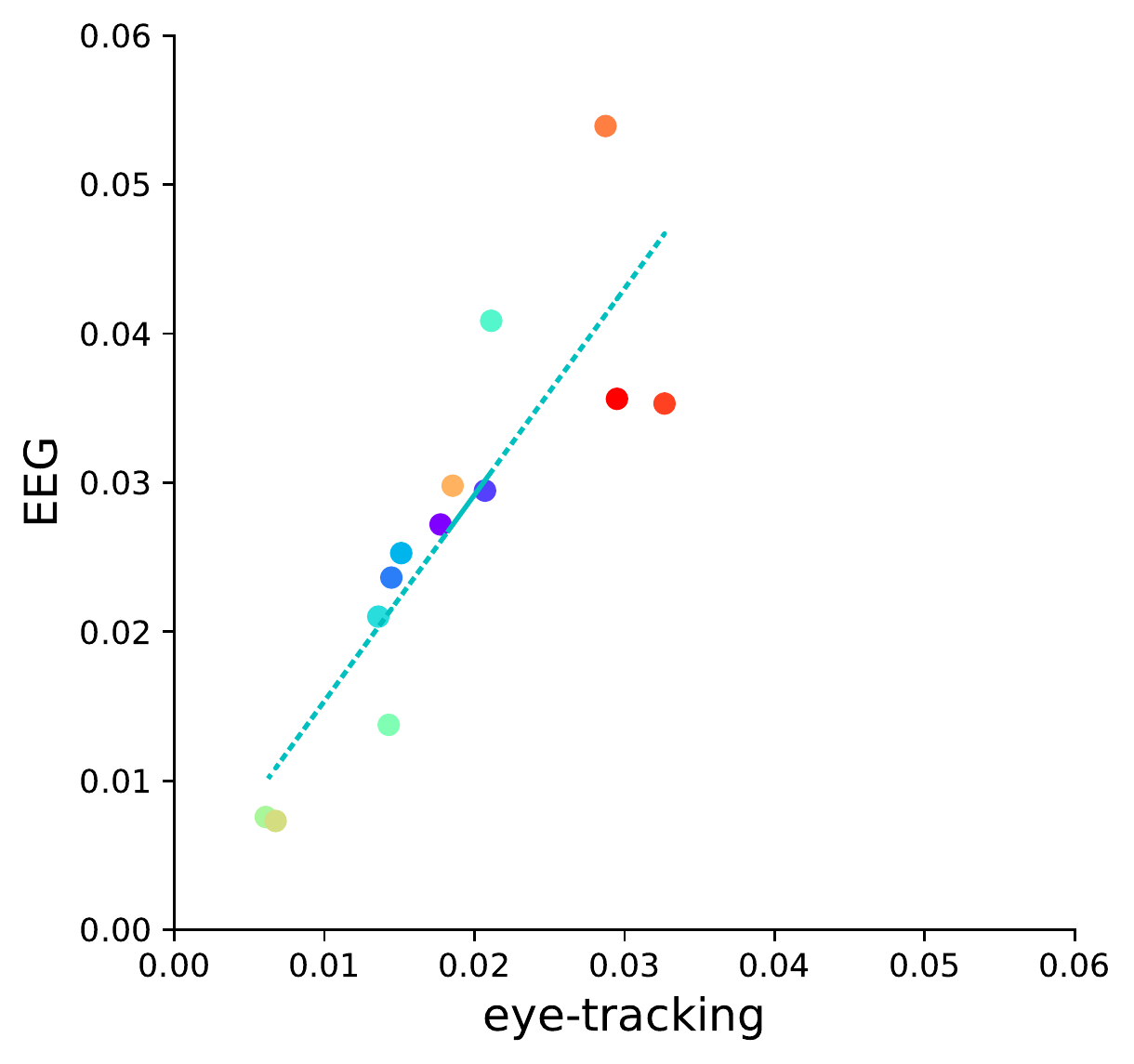}} 
    \subfloat[][]{\includegraphics[width=0.28\textwidth]{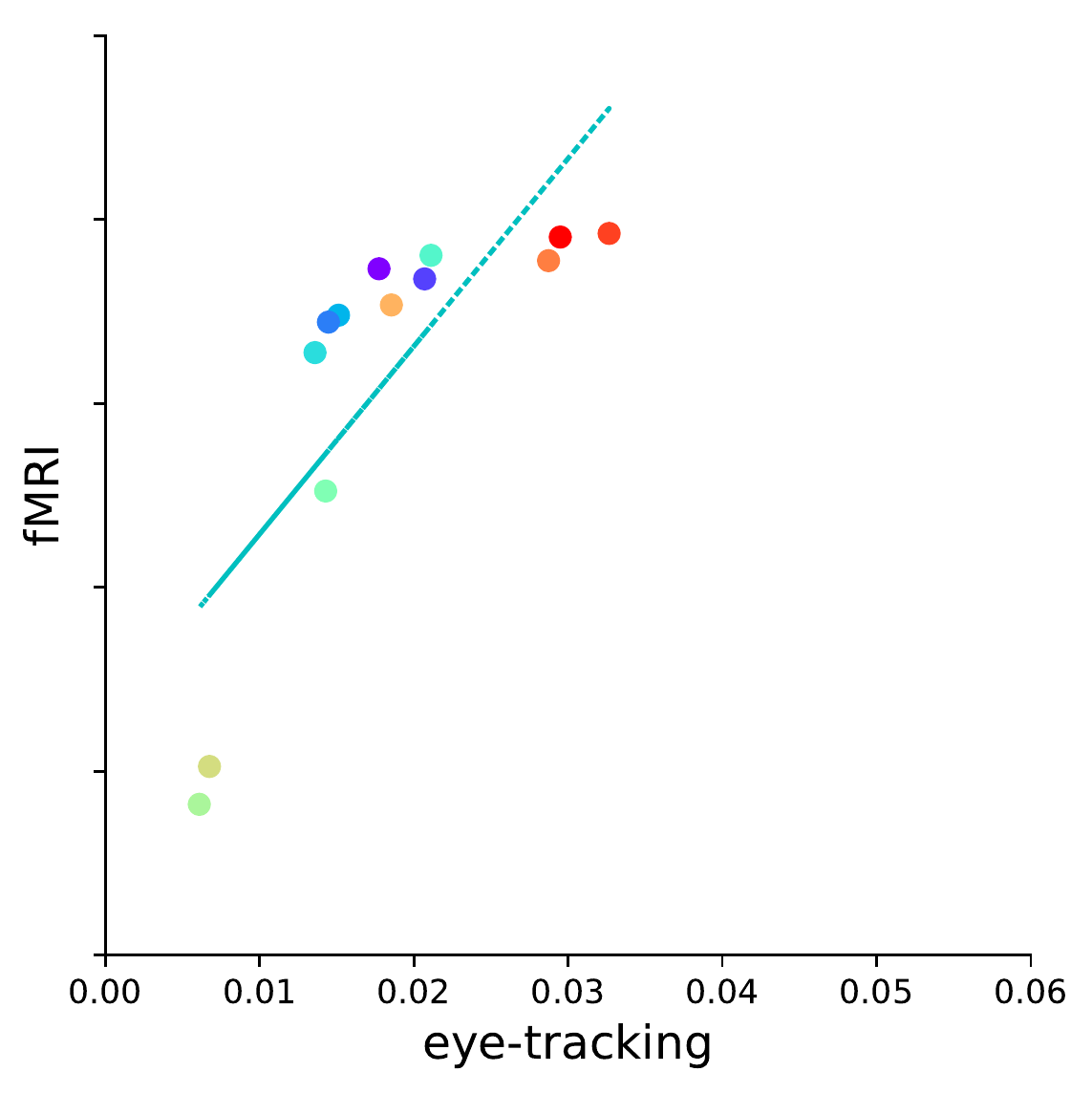}} 
    \subfloat[][]{\includegraphics[width=0.42\textwidth]{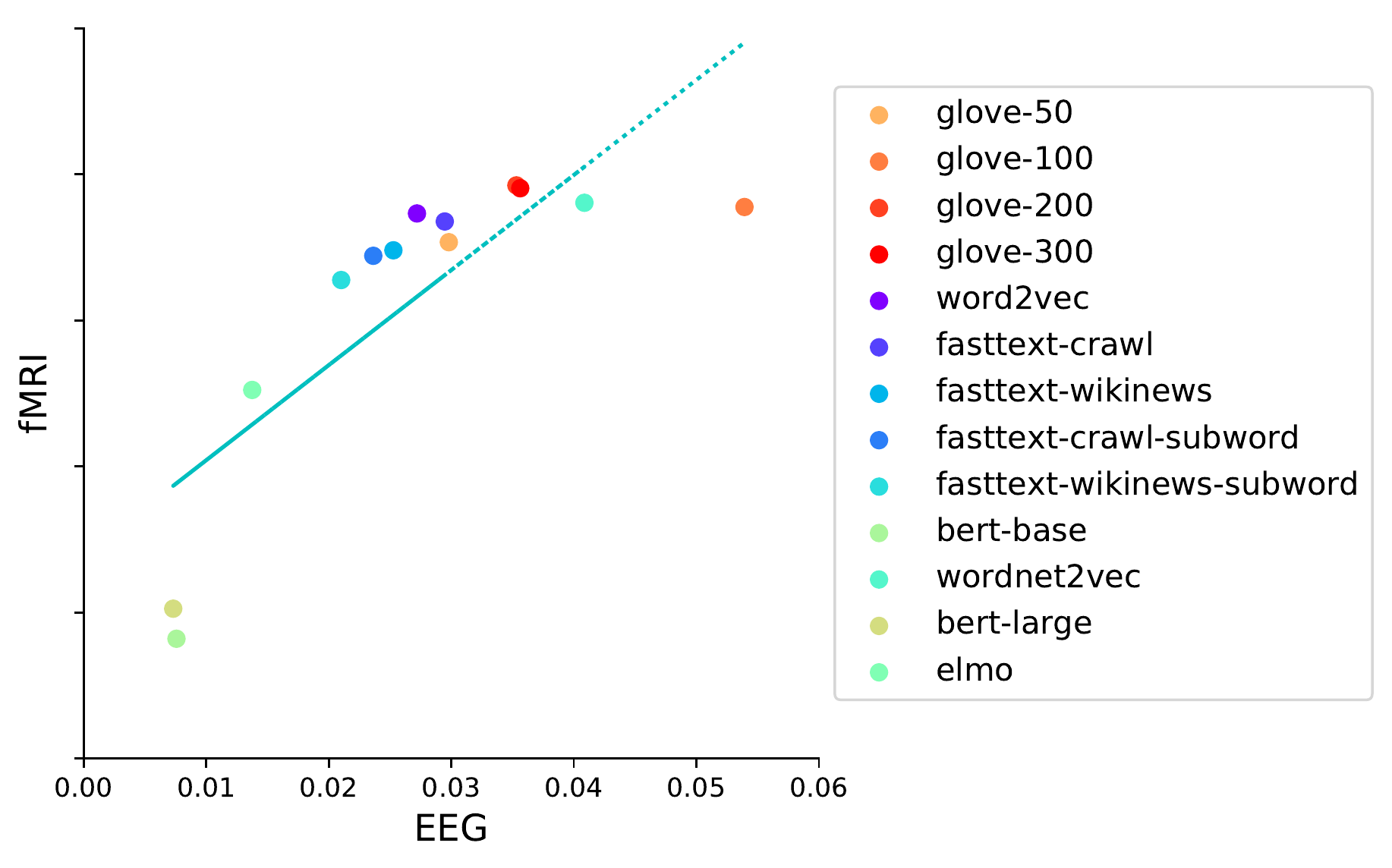}}
    \caption{Correlation plots between all three modalities of cognitive signals.}
    \label{fig:correl-cogni}
\end{figure*}

Since the distribution of our test data is unknown and the datasets are small, we perform a Wilcoxon signed-rank test for each hypothesis \cite{dror2018hitchhiker}. Additionally, to counteract the multiple hypotheses problem, we apply the conservative Bonferroni correction, where the global null hypothesis is rejected if $p < \alpha / N$, where $N$ is the number of hypotheses \cite{dror2017replicability}. In our setting, $\alpha=0.01$ and $N=4$ for EEG (one hypothesis per EEG data source), $N=59$ for for fMRI (one hypothesis per participant of each fMRI data source), and $N=42$ for eye-tracking (one hypothesis per feature per eye-tracking corpus).

This approach of significance testing can easily be used in combination with other intrinsic and extrinsic evaluation methods. The significance ratios are shown in Figure \ref{fig:results-ind}.


\section{Results \& Discussion}

\paragraph{Prediction results} First, we show in Figure \ref{fig:results-ind} how well each word embedding type is able to predict eye-tracking features, EEG and fMRI vectors. As can be seen the majority of results are significantly better than the random baselines. BERT, ELMo and FastText embeddings achieve the best prediction results. All exact numbers can be found in Appendix \ref{app-results}.
While a random baseline can be considered a rather naive choice, this setting also allows us compare the performance between word embedding types.

When predicting single eye-tracking features, the performance varies greatly. For instance, Table \ref{tab:gaze} shows that the prediction error on number of fixations and total reading time from the \textsc{ZuCo} dataset is much lower than for first fixation duration. This suggest that more general eye-tracking features covering the complete reading process of a word are easier to predict. 

In the case of predicting voxel vectors of fMRI data, the results improve when choosing a larger number of voxels (see Table \ref{tab:voxels}). Hence, in the remainder of this work we present only the results for 1000 voxels.

\begin{figure}[h]
    \centering
    \subfloat[][]{\includegraphics[width=.25\textwidth]{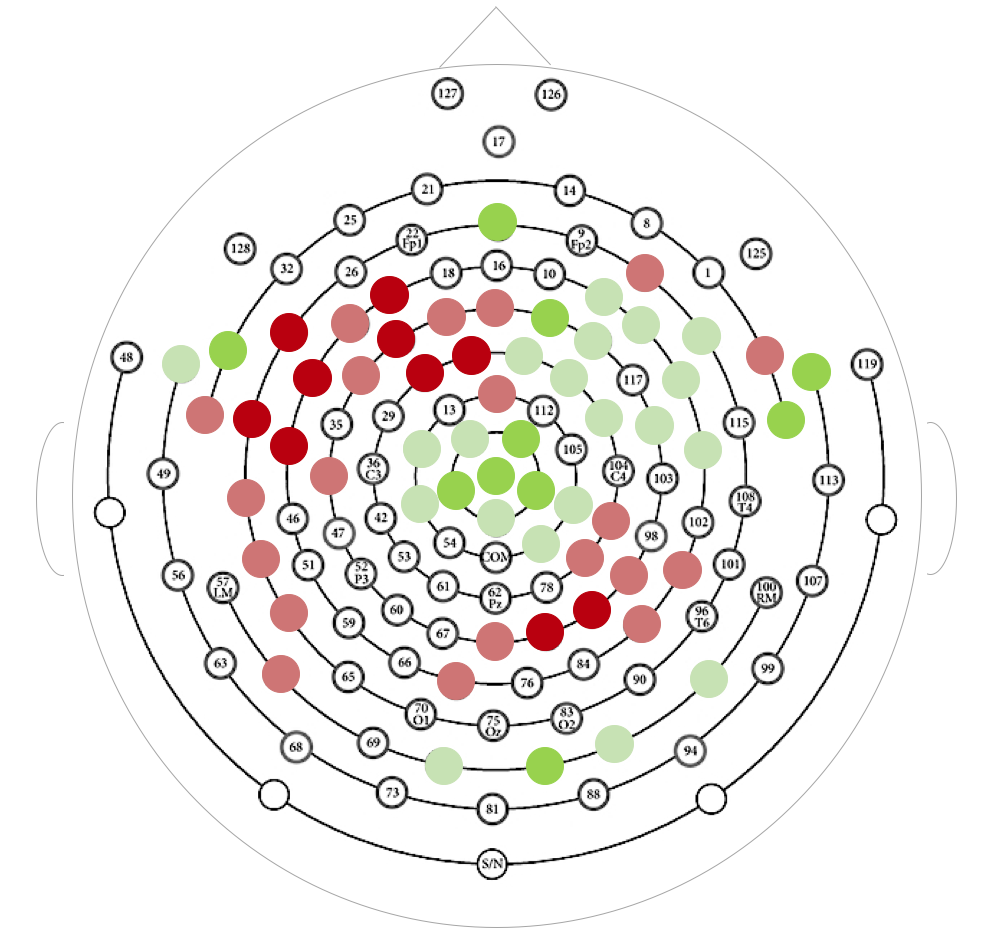}}
   \subfloat[][]{\includegraphics[width=.25\textwidth]{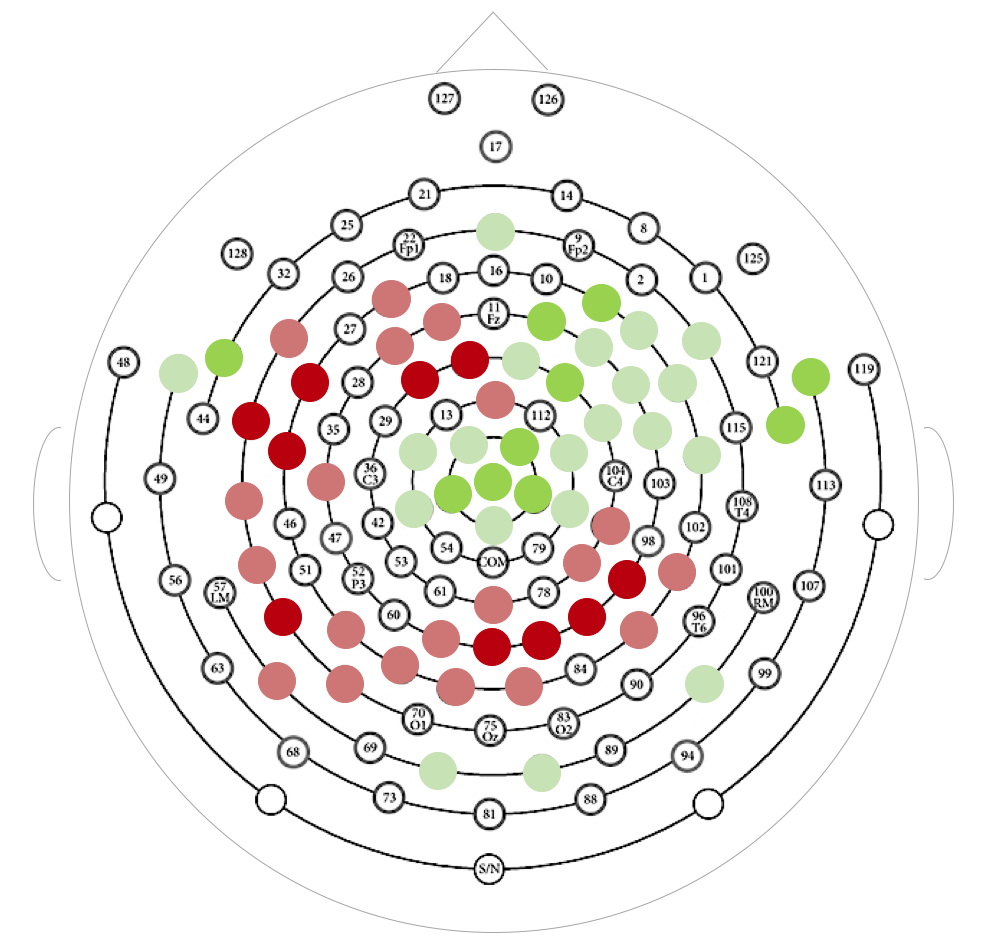}}\\
    \caption{EEG electrode analysis, (a) for BERT (large) and (b) aggregated over all embedding types. Red = worst predicted electrodes, green = best predicted electrodes.}
    \label{fig:electrodes}
\end{figure}

We also examined the EEG results in more depth by analyzing which electrodes are predicted more accurately and which electrodes values are very difficult to predict. This is exemplified by Figure \ref{fig:electrodes}, which shows the 20 best and worst predicted electrodes of the ZuCo data for the BERT embeddings of 1024 dimensions as well as aggregated over all cognitive data sources. The middle central electrodes are predicted more accurately. The middle central electrodes are known to register the activity of the Perisylvian cortex, which is relevant for language related processing \cite{catani2005perisylvian}. Moreover it can be speculated that there is a frontal asymmetry between the electrodes on the left and right hemispheres.

\paragraph{Cognitive data implications} The diversity of cognitive data sources chosen for this work allows us to analyze and compare results on several levels and between several cognitive metrics. In order to conduct this evaluation on a collection of 15 datasets from three modalities, many crucial decisions were taken about preprocessing, feature extraction and evaluation type. Since there are different methods on how to process different types of cognitive language understanding signal, it is important to make these decisions transparent and reproducible.

Moreover, it is a challenge to segment brain activity data correctly and meaningfully into word-level signal from naturalistic, continuous language stimulus \cite{hamilton2018revolution}. This makes consistent preprocessing across data sources even more important.

Another challenge is to consolidate the cognitive features to be predicted. For instance, we chose a wide selection of eye-tracking features that cover early and late word processing. However, choosing only general eye-tracking features such as total reading time would also be a viable strategy. On the other hand, the EEG evaluation could be more coarse-grained, one could also try to predict known ERP effects (e.g. \citet{ettinger2016modeling}) or features selected based on frequency bands. Moreover, the voxel selection in the fMRI preprocessing could be improved by either predicting all voxels or applying information-driven voxel selection methods \cite{beinborn2019robust}.

\begin{figure}[t]
    \centering
    \includegraphics[width=0.48\textwidth]{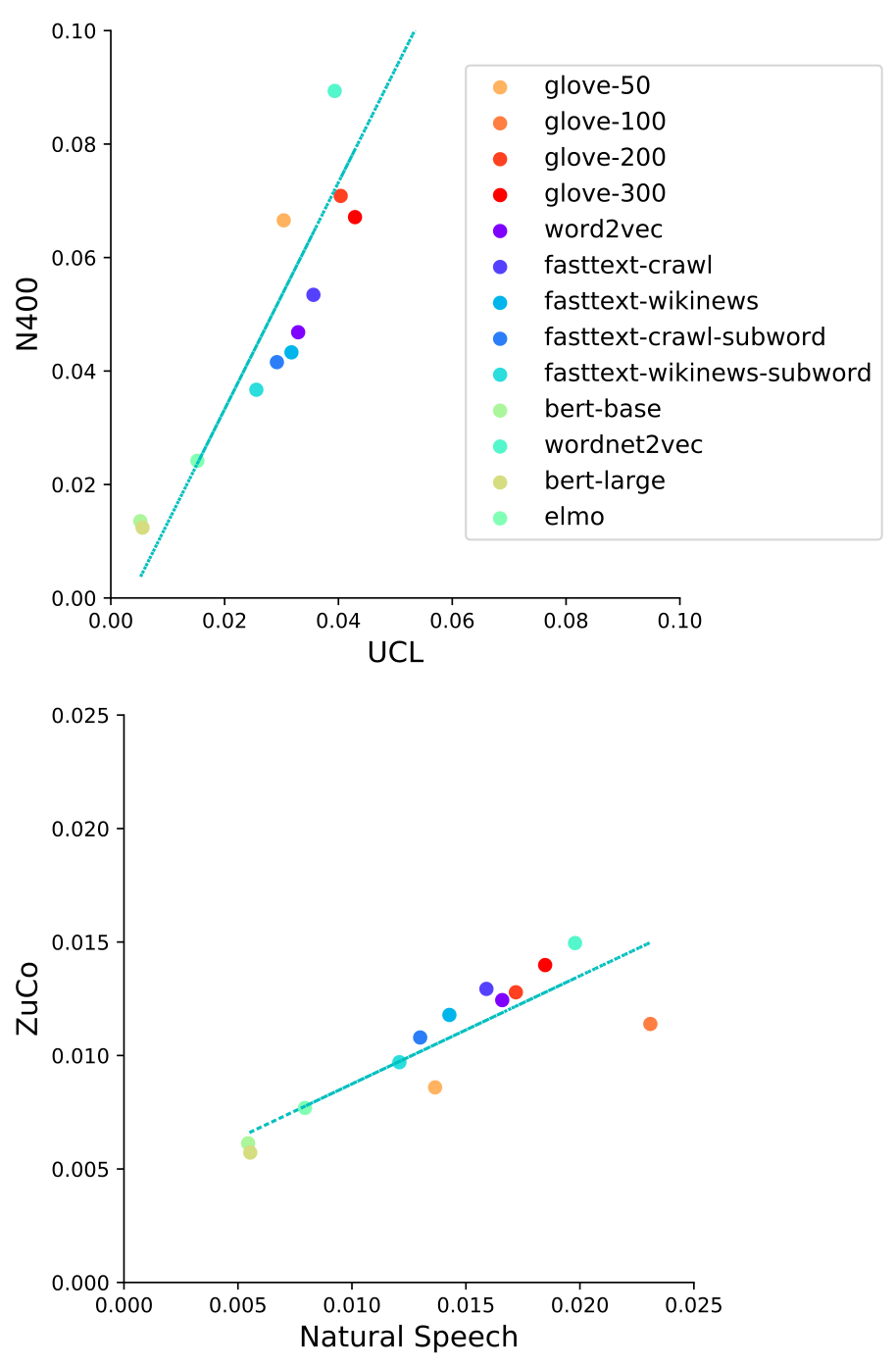}
    \caption{Correlation between results on EEG datasets.}
    \label{fig:correl-eeg}
\end{figure}

\paragraph{Correlations between modalities} Next, we analyze the correlation between the predictions of the three modalities (Figure \ref{fig:correl-cogni}). There is a strong correlation between the results of predicting eye-tracking, EEG and fMRI features. This implies that word embeddings are actually predicting brain activity signals and not merely preprocessing artifacts of each modality. Moreover, the same correlation is also evident between individual datasets within the same modality. As an example, Figure \ref{fig:correl-eeg} (bottom) shows the correlation of the results predicted for the \textsc{Natural Speech} and \textsc{ZuCo} EEG datasets, where the first had speech stimuli and the latter text. Figure \ref{fig:correl-eeg} (top) reveals the same positive correlation for two EEG datasets that were preprocessed differently and were recorded with a different number of electrodes. Moreover, the UCL dataset contains word-by-word reading and the N400 contains natural reading of full sentences.

\begin{figure}[t]
    \centering
    \includegraphics[width=0.4\textwidth]{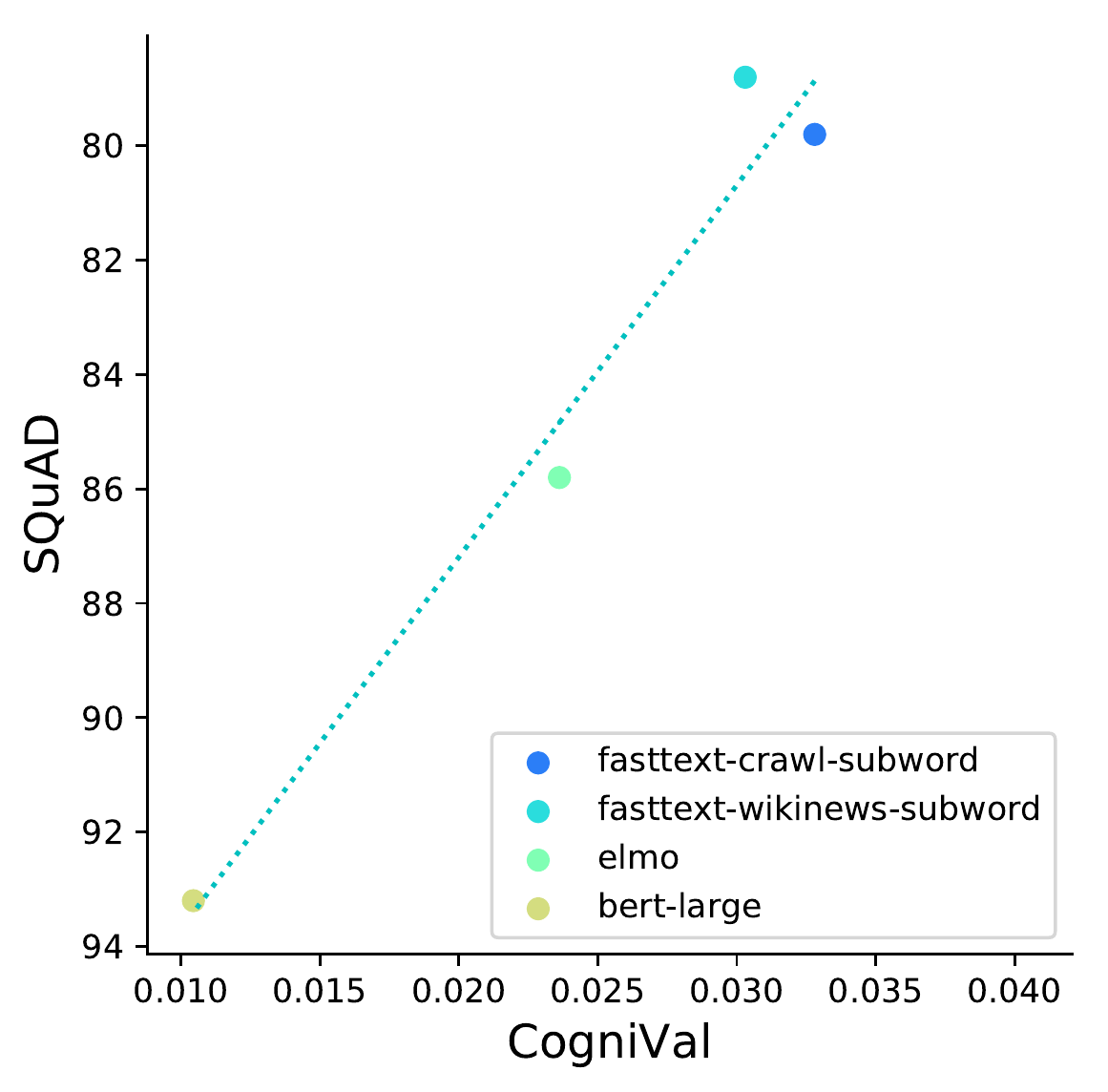}
    \caption{Correlation between the SQuAD 1.1 task and the CogniVal results.}
    \label{fig:squad}
\end{figure}

\begin{figure}[t]
    \centering
    \includegraphics[width=0.4\textwidth]{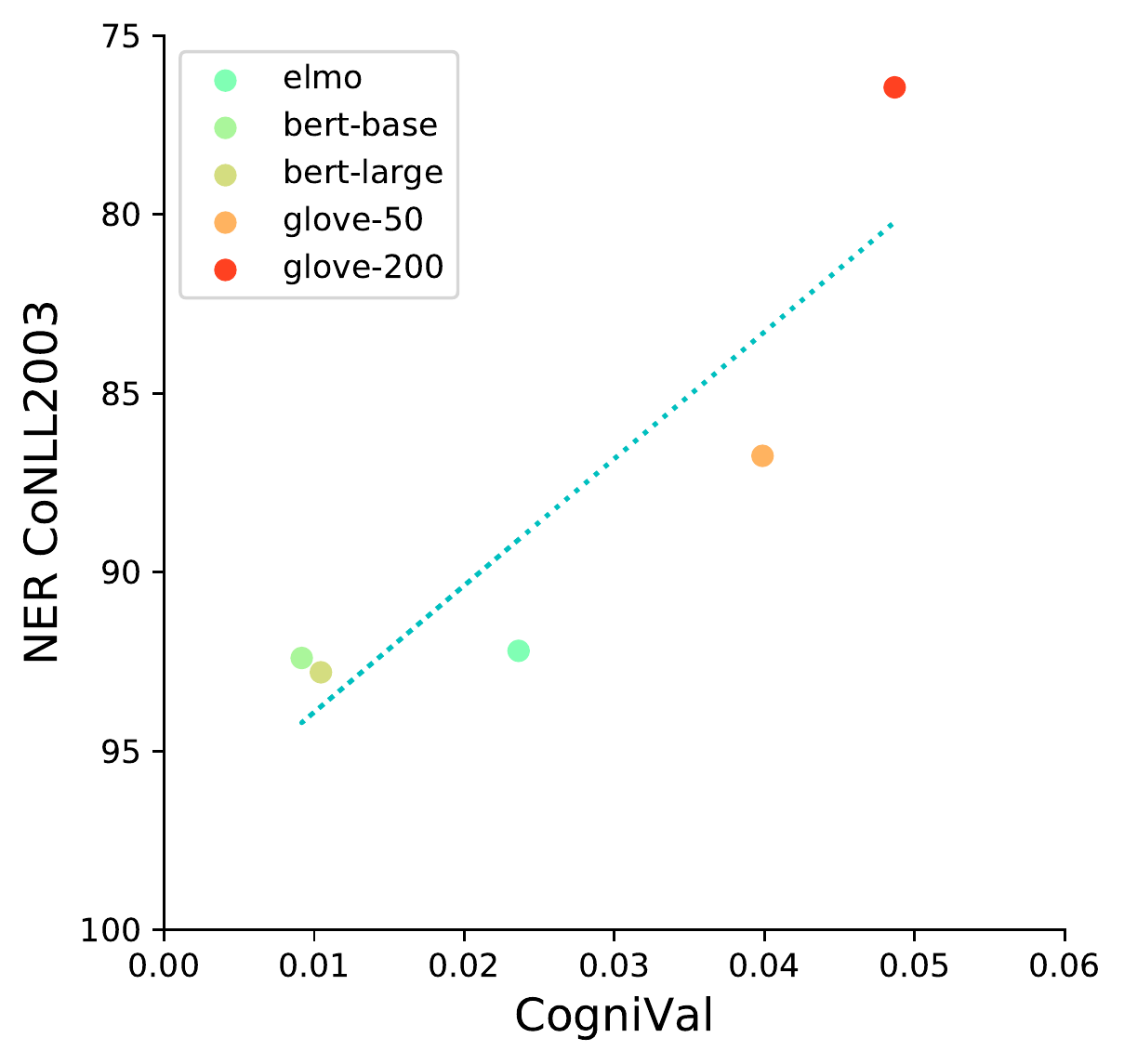}
    \caption{Correlation between NER on CoNLL-2003 and the CogniVal results.}
    \label{fig:ner}
\end{figure}

\paragraph{Correlation with extrinsic evaluation results}
We performed a simple comparison between the results of word embeddings predicting cognitive language processing signals and the performance of the same embedding types in downstream tasks. 
We collected results for two NLP tasks: on the SQuAD 1.1 dataset for question answering \cite{rajpurkar2016squad} and on the CoNLL-2003 test split for named entity recognition \cite{tjong2003introduction}.

The SQuAD results are taken from \citet{devlin2019bert} for BERT, from \citet{mikolov2018advances} for FastText, and from \citet{peters2018deep} for ELMo. The NER results are from the same source for ELMo and BERT, for Glove-50 from \citet{pennington2014glove} and for Glove-200 from \citet{ghannay2016word}. We correlated these results to the prediction results over all cognitive data sources. Figures \ref{fig:squad} and \ref{fig:ner} show the correlation plots between the CogniVal results and the two downstream tasks.

While this is merely an exploratory analysis, it shows interesting findings: If the cognitive embedding evaluation correlates with the performance of the embeddings in extrinsic evaluation tasks, it might be used not only for evaluation but also as a predictive framework for word embedding model selection. This is especially noteworthy, since it does not seem to be the case for other intrinsic methods \cite{chiu2016intrinsic}.

\section{Conclusion}

We presented CogniVal, the first multi-modal large-scale cognitive word embedding evaluation framework. The vectorized word representations are evaluated by using them to predict eye-tracking or brain activity data recorded while participants were understanding natural language. We find that the results of eye-tracking, EEG and fMRI data are strongly correlated not only across these modalities but even between datasets within the same modality. Intriguinly, we also find a correlation between our cognitive evaluation and two extrinsic NLP tasks, which opens the question whether CogniVal can also be used for predicting downstream performance and hence, choosing the best embeddings for specific tasks.

We plan to expand the collection of cognitive data sources as more of them become available, including data from other languages such as the Narrative Brain Dataset (Dutch, fMRI, \citet{lopopolo2018narrative}) or the Russian Sentence Corpus (eye-tracking, \citet{laurinavichyute2017russian}). Thanks to naturalistic recording of longer text spans, CogniVal can also be extended to evaluate sentence embeddings or even paragraph embeddings.

CogniVal can become even more effective by combining the results with other intrinsic or extrinsic embedding evaluation frameworks \cite{nayak2016evaluating,rogers2018what} and building on the multiple hypotheses testing.


\section{Acknowledgements}
We thank Lisa Beinborn, Stefan Frank and Thomas Lemmin for their valuable input on preprocessing EEG and fMRI data.

\bibliography{emnlp-ijcnlp-2019.bib}

\begin{thebibliography}{53}
\expandafter\ifx\csname natexlab\endcsname\relax\def\natexlab#1{#1}\fi

\bibitem[{Abnar et~al.(2018)Abnar, Ahmed, Mijnheer, and
  Zuidema}]{abnar2018experiential}
Samira Abnar, Rasyan Ahmed, Max Mijnheer, and Willem Zuidema. 2018.
\newblock Experiential, distributional and dependency-based word embeddings
  have complementary roles in decoding brain activity.
\newblock In \emph{Proceedings of the 8th Workshop on Cognitive Modeling and
  Computational Linguistics (CMCL 2018)}, pages 57--66.

\bibitem[{Bakarov(2018{\natexlab{a}})}]{bakarov2018can}
Amir Bakarov. 2018{\natexlab{a}}.
\newblock Can eye movement data be used as ground truth for word embeddings
  evaluation?
\newblock \emph{arXiv preprint arXiv:1804.08749}.

\bibitem[{Bakarov(2018{\natexlab{b}})}]{bakarov2018survey}
Amir Bakarov. 2018{\natexlab{b}}.
\newblock A survey of word embeddings evaluation methods.
\newblock \emph{arXiv preprint arXiv:1801.09536}.

\bibitem[{Beinborn et~al.(2019)Beinborn, Abnar, and
  Choenni}]{beinborn2019robust}
Lisa Beinborn, Samira Abnar, and Rochelle Choenni. 2019.
\newblock Robust evaluation of language-brain encoding experiments.
\newblock \emph{arXiv preprint arXiv:1904.02547}.

\bibitem[{Brennan et~al.(2016)Brennan, Stabler, Van~Wagenen, Luh, and
  Hale}]{brennan2016abstract}
Jonathan~R Brennan, Edward~P Stabler, Sarah~E Van~Wagenen, Wen-Ming Luh, and
  John~T Hale. 2016.
\newblock Abstract linguistic structure correlates with temporal activity
  during naturalistic comprehension.
\newblock \emph{Brain and Language}, 157:81--94.

\bibitem[{Broderick et~al.(2018)Broderick, Anderson, Di~Liberto, Crosse, and
  Lalor}]{broderick2018electrophysiological}
Michael~P Broderick, Andrew~J Anderson, Giovanni~M Di~Liberto, Michael~J
  Crosse, and Edmund~C Lalor. 2018.
\newblock Electrophysiological correlates of semantic dissimilarity reflect the
  comprehension of natural, narrative speech.
\newblock \emph{Current Biology}, 28(5):803--809.

\bibitem[{Catani et~al.(2005)Catani, Jones, and
  Ffytche}]{catani2005perisylvian}
Marco Catani, Derek~K Jones, and Dominic~H Ffytche. 2005.
\newblock Perisylvian language networks of the human brain.
\newblock \emph{Annals of Neurology: Official Journal of the American
  Neurological Association and the Child Neurology Society}, 57(1):8--16.

\bibitem[{Chiu et~al.(2016)Chiu, Korhonen, and Pyysalo}]{chiu2016intrinsic}
Billy Chiu, Anna Korhonen, and Sampo Pyysalo. 2016.
\newblock Intrinsic evaluation of word vectors fails to predict extrinsic
  performance.
\newblock In \emph{Proceedings of the 1st Workshop on Evaluating Vector-Space
  Representations for NLP}, pages 1--6.

\bibitem[{Cop et~al.(2017)Cop, Dirix, Drieghe, and Duyck}]{cop2017presenting}
Uschi Cop, Nicolas Dirix, Denis Drieghe, and Wouter Duyck. 2017.
\newblock Presenting {GECO}: An eyetracking corpus of monolingual and bilingual
  sentence reading.
\newblock \emph{Behavior research methods}, 49(2):602--615.

\bibitem[{Devlin et~al.(2019)Devlin, Chang, Lee, and
  Toutanova}]{devlin2019bert}
Jacob Devlin, Ming-Wei Chang, Kenton Lee, and Kristina Toutanova. 2019.
\newblock {BERT}: Pre-training of deep bidirectional transformers for language
  understanding.
\newblock In \emph{Proceedings of the 2019 Conference of the North American
  Chapter of the Association for Computational Linguistics: Human Language
  Technologies, Volume 1 (Long and Short Papers)}, pages 4171--4186.

\bibitem[{Dror et~al.(2017)Dror, Baumer, Bogomolov, and
  Reichart}]{dror2017replicability}
Rotem Dror, Gili Baumer, Marina Bogomolov, and Roi Reichart. 2017.
\newblock Replicability analysis for natural language processing: Testing
  significance with multiple datasets.
\newblock \emph{Transactions of the Association for Computational Linguistics},
  5:471--486.

\bibitem[{Dror et~al.(2018)Dror, Baumer, Shlomov, and
  Reichart}]{dror2018hitchhiker}
Rotem Dror, Gili Baumer, Segev Shlomov, and Roi Reichart. 2018.
\newblock The hitchhiker's guide to testing statistical significance in natural
  language processing.
\newblock In \emph{Proceedings of the 56th Annual Meeting of the Association
  for Computational Linguistics (Volume 1: Long Papers)}, pages 1383--1392.

\bibitem[{Ettinger et~al.(2016)Ettinger, Feldman, Resnik, and
  Phillips}]{ettinger2016modeling}
Allyson Ettinger, Naomi Feldman, Philip Resnik, and Colin Phillips. 2016.
\newblock Modeling {N400} amplitude using vector space models of word
  representation.
\newblock In \emph{CogSci}.

\bibitem[{Frank(2017)}]{frank2017word}
Stefan~L Frank. 2017.
\newblock Word embedding distance does not predict word reading time.

\bibitem[{Frank et~al.(2013)Frank, Monsalve, Thompson, and
  Vigliocco}]{frank2013reading}
Stefan~L Frank, Irene~Fernandez Monsalve, Robin~L Thompson, and Gabriella
  Vigliocco. 2013.
\newblock Reading time data for evaluating broad-coverage models of english
  sentence processing.
\newblock \emph{Behavior Research Methods}, 45(4):1182--1190.

\bibitem[{Frank et~al.(2015)Frank, Otten, Galli, and Vigliocco}]{frank2015erp}
Stefan~L Frank, Leun~J Otten, Giulia Galli, and Gabriella Vigliocco. 2015.
\newblock The {ERP} response to the amount of information conveyed by words in
  sentences.
\newblock \emph{Brain and language}, 140:1--11.

\bibitem[{Frank and Willems(2017)}]{frank2017wordpred}
Stefan~L Frank and Roel~M Willems. 2017.
\newblock Word predictability and semantic similarity show distinct patterns of
  brain activity during language comprehension.
\newblock \emph{Language, Cognition and Neuroscience}, 32(9):1192--1203.

\bibitem[{Ghannay et~al.(2016)Ghannay, Favre, Esteve, and
  Camelin}]{ghannay2016word}
Sahar Ghannay, Benoit Favre, Yannick Esteve, and Nathalie Camelin. 2016.
\newblock Word embedding evaluation and combination.
\newblock In \emph{Proceedings of the Tenth International Conference on
  Language Resources and Evaluation (LREC 2016)}, pages 300--305.

\bibitem[{Gladkova and Drozd(2016)}]{gladkova2016intrinsic}
Anna Gladkova and Aleksandr Drozd. 2016.
\newblock Intrinsic evaluations of word embeddings: What can we do better?
\newblock In \emph{Proceedings of the 1st Workshop on Evaluating Vector-Space
  Representations for NLP}, pages 36--42.

\bibitem[{Hamilton and Huth(2018)}]{hamilton2018revolution}
Liberty~S Hamilton and Alexander~G Huth. 2018.
\newblock The revolution will not be controlled: Natural stimuli in speech
  neuroscience.
\newblock \emph{Language, Cognition and Neuroscience}, pages 1--10.

\bibitem[{Hauk and Pulverm{\"u}ller(2004)}]{hauk2004effects}
Olaf Hauk and Friedemann Pulverm{\"u}ller. 2004.
\newblock Effects of word length and frequency on the human event-related
  potential.
\newblock \emph{Clinical Neurophysiology}, 115(5):1090--1103.

\bibitem[{Hollenstein et~al.(2018)Hollenstein, Rotsztejn, Troendle, Pedroni,
  Zhang, and Langer}]{hollenstein2018zuco}
Nora Hollenstein, Jonathan Rotsztejn, Marius Troendle, Andreas Pedroni,
  Ce~Zhang, and Nicolas Langer. 2018.
\newblock {ZuCo, a simultaneous EEG} and eye-tracking resource for natural
  sentence reading.
\newblock \emph{Scientific Data}.

\bibitem[{Hollenstein and Zhang(2019)}]{hollenstein2019entity}
Nora Hollenstein and Ce~Zhang. 2019.
\newblock Entity recognition at first sight: Improving {NER} with eye movement
  information.
\newblock In \emph{NAACL}.

\bibitem[{Huth et~al.(2016)Huth, de~Heer, Griffiths, Theunissen, and
  Gallant}]{huth2016natural}
Alexander~G Huth, Wendy~A de~Heer, Thomas~L Griffiths, Fr{\'e}d{\'e}ric~E
  Theunissen, and Jack~L Gallant. 2016.
\newblock Natural speech reveals the semantic maps that tile human cerebral
  cortex.
\newblock \emph{Nature}, 532(7600):453--458.

\bibitem[{Just and Carpenter(1980)}]{just1980theory}
Marcel~A Just and Patricia~A Carpenter. 1980.
\newblock A theory of reading: From eye fixations to comprehension.
\newblock \emph{Psychological review}, 87(4):329.

\bibitem[{Kennedy et~al.(2003)Kennedy, Hill, and Pynte}]{kennedy2003dundee}
Alan Kennedy, Robin Hill, and Jo{\"e}l Pynte. 2003.
\newblock The {Dundee} corpus.
\newblock In \emph{Proceedings of the 12th European Conference on Eye
  Movement}.

\bibitem[{Laurinavichyute et~al.(2017)Laurinavichyute, Sekerina, Alexeeva, and
  Bagdasaryan}]{laurinavichyute2017russian}
AK~Laurinavichyute, Irina~A Sekerina, SV~Alexeeva, and KA~Bagdasaryan. 2017.
\newblock {Russian Sentence Corpus}: Benchmark measures of eye movements in
  reading in {C}yrillic.

\bibitem[{Lopopolo et~al.(2018)Lopopolo, Frank, Van~den Bosch, Nijhof, and
  Willems}]{lopopolo2018narrative}
Alessandro Lopopolo, Stefan~L Frank, Antal Van~den Bosch, Annabel Nijhof, and
  Roel~M Willems. 2018.
\newblock The {Narrative Brain Dataset (NBD)}, an {fMRI} dataset for the study
  of natural language processing in the brain.
\newblock In \emph{LREC 2018 Workshop on Linguistic and Neuro-Cognitive
  Resources (LiNCR)}. LREC.

\bibitem[{Luke and Christianson(2017)}]{luke2017provo}
Steven~G Luke and Kiel Christianson. 2017.
\newblock The {Provo Corpus}: A large eye-tracking corpus with predictability
  norms.
\newblock \emph{Behavior Research Methods}, pages 1--8.

\bibitem[{Malvina~Nissim(2019)}]{nissim2019fair}
Rob van der~Goot Malvina~Nissim, Rik van~Noord. 2019.
\newblock Fair is better than sensational: Man is to doctor as woman is to
  doctor.
\newblock \emph{arXiv preprint arXiv:1905.09866}.

\bibitem[{Miezin et~al.(2000)Miezin, Maccotta, Ollinger, Petersen, and
  Buckner}]{miezin2000characterizing}
Francis~M Miezin, L~Maccotta, JM~Ollinger, SE~Petersen, and RL~Buckner. 2000.
\newblock Characterizing the hemodynamic response: effects of presentation
  rate, sampling procedure, and the possibility of ordering brain activity
  based on relative timing.
\newblock \emph{Neuroimage}, 11(6):735--759.

\bibitem[{Mikolov et~al.(2018)Mikolov, Grave, Bojanowski, Puhrsch, and
  Joulin}]{mikolov2018advances}
Tomas Mikolov, Edouard Grave, Piotr Bojanowski, Christian Puhrsch, and Armand
  Joulin. 2018.
\newblock Advances in pre-training distributed word representations.
\newblock In \emph{Proceedings of the International Conference on Language
  Resources and Evaluation (LREC 2018)}.

\bibitem[{Mikolov et~al.(2013)Mikolov, Sutskever, Chen, Corrado, and
  Dean}]{mikolov2013distributed}
Tomas Mikolov, Ilya Sutskever, Kai Chen, Greg~S Corrado, and Jeff Dean. 2013.
\newblock Distributed representations of words and phrases and their
  compositionality.
\newblock In \emph{Advances in neural information processing systems}, pages
  3111--3119.

\bibitem[{Miller and Fellbaum(1992)}]{miller1992wordnet}
George~A Miller and Christiane Fellbaum. 1992.
\newblock Wordnet and the organization of lexical memory.
\newblock In \emph{Intelligent tutoring systems for foreign language learning},
  pages 89--102. Springer.

\bibitem[{Mishra et~al.(2016)Mishra, Kanojia, and
  Bhattacharyya}]{mishra2016predicting}
Abhijit Mishra, Diptesh Kanojia, and Pushpak Bhattacharyya. 2016.
\newblock Predicting readers' sarcasm understandability by modeling gaze
  behavior.
\newblock In \emph{AAAI}, pages 3747--3753.

\bibitem[{Mishra et~al.(2017)Mishra, Kanojia, Nagar, Dey, and
  Bhattacharyya}]{mishra2017scanpath}
Abhijit Mishra, Diptesh Kanojia, Seema Nagar, Kuntal Dey, and Pushpak
  Bhattacharyya. 2017.
\newblock Scanpath complexity: Modeling reading effort using gaze information.
\newblock In \emph{AAAI}, pages 4429--4436.

\bibitem[{Mitchell et~al.(2008)Mitchell, Shinkareva, Carlson, Chang, Malave,
  Mason, and Just}]{mitchell2008predicting}
Tom~M Mitchell, Svetlana~V Shinkareva, Andrew Carlson, Kai-Min Chang, Vicente~L
  Malave, Robert~A Mason, and Marcel~Adam Just. 2008.
\newblock Predicting human brain activity associated with the meanings of
  nouns.
\newblock \emph{Science}, 320(5880):1191--1195.

\bibitem[{Mulert(2013)}]{mulert2013simultaneous}
Christoph Mulert. 2013.
\newblock Simultaneous {EEG} and {fMRI}: towards the characterization of
  structure and dynamics of brain networks.
\newblock \emph{Dialogues in clinical neuroscience}, 15(3):381.

\bibitem[{Murphy et~al.(2018)Murphy, Wehbe, and Fyshe}]{murphy2018decoding}
Brian Murphy, Leila Wehbe, and Alona Fyshe. 2018.
\newblock Decoding language from the brain.
\newblock \emph{Language, cognition, and computational models}, page~53.

\bibitem[{Nayak et~al.(2016)Nayak, Angeli, and Manning}]{nayak2016evaluating}
Neha Nayak, Gabor Angeli, and Christopher~D Manning. 2016.
\newblock Evaluating word embeddings using a representative suite of practical
  tasks.
\newblock In \emph{Proceedings of the 1st Workshop on Evaluating Vector-Space
  Representations for NLP}, pages 19--23.

\bibitem[{Pennington et~al.(2014)Pennington, Socher, and
  Manning}]{pennington2014glove}
Jeffrey Pennington, Richard Socher, and Christopher~D. Manning. 2014.
\newblock \href {http://www.aclweb.org/anthology/D14-1162} {Glove: Global
  vectors for word representation}.
\newblock In \emph{Proceedings of the 2014 Conference on Empirical Methods in
  Natural Language Processing}, pages 1532--1543.

\bibitem[{Pereira et~al.(2018)Pereira, Lou, Pritchett, Ritter, Gershman,
  Kanwisher, Botvinick, and Fedorenko}]{pereira2018toward}
Francisco Pereira, Bin Lou, Brianna Pritchett, Samuel Ritter, Samuel~J
  Gershman, Nancy Kanwisher, Matthew Botvinick, and Evelina Fedorenko. 2018.
\newblock Toward a universal decoder of linguistic meaning from brain
  activation.
\newblock \emph{Nature communications}, 9(1):963.

\bibitem[{Peters et~al.(2018)Peters, Neumann, Iyyer, Gardner, Clark, Lee, and
  Zettlemoyer}]{peters2018deep}
Matthew~E. Peters, Mark Neumann, Mohit Iyyer, Matt Gardner, Christopher Clark,
  Kenton Lee, and Luke Zettlemoyer. 2018.
\newblock Deep contextualized word representations.
\newblock In \emph{Proceedings of NAACL}.

\bibitem[{Price(2012)}]{price2012review}
Cathy~J Price. 2012.
\newblock A review and synthesis of the first 20 years of {PET} and {fMRI}
  studies of heard speech, spoken language and reading.
\newblock \emph{Neuroimage}, 62(2):816--847.

\bibitem[{Rajpurkar et~al.(2016)Rajpurkar, Zhang, Lopyrev, and
  Liang}]{rajpurkar2016squad}
Pranav Rajpurkar, Jian Zhang, Konstantin Lopyrev, and Percy Liang. 2016.
\newblock {SQuAD: 100,000+} questions for machine comprehension of text.
\newblock In \emph{Proceedings of the 2016 Conference on Empirical Methods in
  Natural Language Processing}, pages 2383--2392.

\bibitem[{Rayner(1998)}]{rayner1998eye}
Keith Rayner. 1998.
\newblock Eye movements in reading and information processing: 20 years of
  research.
\newblock \emph{Psychological bulletin}, 124(3):372.

\bibitem[{Rodrigues et~al.(2018)Rodrigues, Branco, Silva, Saedi, and
  Branco}]{rodrigues2018predicting}
Joao~Ant{\'o}nio Rodrigues, Ruben Branco, Jo{\~a}o Silva, Chakaveh Saedi, and
  Ant{\'o}nio Branco. 2018.
\newblock Predicting brain activation with {WordNet} embeddings.
\newblock In \emph{Proceedings of the Eight Workshop on Cognitive Aspects of
  Computational Language Learning and Processing}, pages 1--5.

\bibitem[{Rogers et~al.(2018)Rogers, Ananthakrishna, and
  Rumshisky}]{rogers2018what}
Anna Rogers, Shashwath~Hosur Ananthakrishna, and Anna Rumshisky. 2018.
\newblock What's in your embedding, and how it predicts task performance.
\newblock In \emph{Proceedings of the 27th International Conference on
  Computational Linguistics}, pages 2690--2703.

\bibitem[{Saedi et~al.(2018)Saedi, Branco, Rodrigues, and
  Silva}]{saedi2018wordnet}
Chakaveh Saedi, Ant{\'o}nio Branco, Jo{\~a}o~Ant{\'o}nio Rodrigues, and
  Jo{\~a}o Silva. 2018.
\newblock {WordNet} embeddings.
\newblock In \emph{Proceedings of The Third Workshop on Representation Learning
  for NLP}, pages 122--131.

\bibitem[{Schwartz and Mitchell(2019)}]{schwartz2019understanding}
Dan Schwartz and Tom Mitchell. 2019.
\newblock Understanding language-elicited {EEG} data by predicting it from a
  fine-tuned language model.
\newblock In \emph{Proceedings of the 2019 Conference of the North American
  Chapter of the Association for Computational Linguistics: Human Language
  Technologies, Volume 1 (Long and Short Papers)}, pages 43--57.

\bibitem[{S{\o}gaard(2016)}]{sogaard2016evaluating}
Anders S{\o}gaard. 2016.
\newblock {Evaluating word embeddings with fMRI and eye-tracking}.
\newblock In \emph{Proceedings of the 1st Workshop on Evaluating Vector-Space
  Representations for NLP}, pages 116--121.

\bibitem[{Tjong Kim~Sang and De~Meulder(2003)}]{tjong2003introduction}
Erik~F Tjong Kim~Sang and Fien De~Meulder. 2003.
\newblock Introduction to the {CoNLL-2003} shared task: Language-independent
  named entity recognition.
\newblock In \emph{Proceedings of the 7th Conference on Natural Language
  Learning}, volume~4, pages 142--147.

\bibitem[{Wehbe et~al.(2014)Wehbe, Vaswani, Knight, and
  Mitchell}]{wehbe2014aligning}
Leila Wehbe, Ashish Vaswani, Kevin Knight, and Tom Mitchell. 2014.
\newblock Aligning context-based statistical models of language with brain
  activity during reading.
\newblock In \emph{Proceedings of the 2014 Conference on Empirical Methods in
  Natural Language Processing (EMNLP)}, pages 233--243.

\end{thebibliography}
\bibliographystyle{acl_natbib}

\newpage
\onecolumn
\appendix

\section{Data Preprocessing}

\subsection{Eye-tracking Features}\label{app-gaze}

\begin{table*}[h]
\centering
\begin{tabular}{p{9.3cm}p{5.4cm}}
\hline
\textbf{Description} & \textbf{Data source} \\\hline\hline
First fixation duration & Dundee, GECO, Provo, UCL, ZuCo \\\hline
First pass duration (first fixation duration in the first pass reading) & Dundee, GECO, Provo, UCL, ZuCo \\\hline
Mean fixation duration & Dundee, GECO, Provo, ZuCo, CFILT-Sarcasm, CFILT-Scanpath \\\hline
Fixation probability & Dundee \\\hline
Re-read probability & Dundee \\\hline
Total fixation duration & Dundee, GECO, Provo, ZuCo \\\hline
Total duration of all regression going from this word & Dundee \\\hline
Total duration of all regression going to this word & Dundee \\\hline
Number of fixations & Dundee, GECO, Provo, ZuCo \\\hline
Number of long regression (\textgreater 3 tokens) going from this word & Dundee \\\hline
Number of long regression (\textgreater 3 tokens) going to this word & Dundee \\\hline
Number of refixations & Dundee \\\hline
Number of regressions going from this word & Dundee, Provo \\\hline
Number of regressions going to this word & Dundee, Provo \\\hline
The duration of the last fixation on the current word & GECO \\\hline
Go-past time & GECO, Provo, UCL, ZuCo \\\hline
No fixation occurred in first-pass reading & GECO, Provo \\\hline
Right-bounded reading time & UCL \\\hline
\end{tabular}
\caption{Eye-tracking features provded in the gaze corpora used in this work}
\label{tab:et-features}
\end{table*}

\subsection{EEG}\label{app-eeg}

All four EEG datasets are converted to the EEGLab format\footnote{https://sccn.ucsd.edu/eeglab/index.php}, if not already provided in this format. The UCL dataset had been preprocessed by the authors.
For the other three datasets, bandpass filtering, artifact removal (i.e. removing blinks and other muscle activity) and quality assessment was performed with Automagic\footnote{https://github.com/methlabUZH/automagic}. 

After preprocessing and retaining only the subjects with good data quality, we use the data of 3 subjects from the N400 dataset, 14 subjects from Natural Speech, 12 subject from ZuCo and the same number of subjects as originally from UCL (i.e. 24).

\subsection{fMRI}\label{app-fmri}

As mentioned in the main paper, we use the preprocessing pipeline from \citet{beinborn2019robust} to read the fMRI data, align the scans and select the voxels. We used the \textsc{Nouns} and \textsc{Pereira} readers as is and modified the \textsc{Harry Potter} and \textsc{Alice} readers to extract word-level signals.

\newpage

\section{Detailed Results}\label{app-results}

\begin{table}[h]
\centering
\begin{tabular}{lccccccc}
\hline
\textbf{embeddings} & \textsc{GECO} & \textsc{ZuCo} & \textsc{Provo} & \textsc{Dundee} & \textsc{Sarcasm} & \textsc{Scanpath} & \textsc{UCL} \\\hline\hline
glove-50 & 0.010 & 0.008 & 0.031 & 0.010 & 0.016 & 0.023 & 0.044 \\
glove-100 & 0.018 & 0.017 & 0.051 & 0.014 & 0.027 & 0.043 & 0.054 \\
glove-200 & 0.026 & 0.024 & 0.047 & 0.021 & 0.039 & 0.038 & 0.054 \\
glove-300 & 0.020 & 0.019 & 0.047 & 0.016 & 0.033 & 0.038 & 0.059 \\
word2vec & 0.015 & 0.011 & 0.024 & 0.014 & 0.022 & 0.018 & 0.028 \\
fasttext-crawl & 0.011 & 0.009 & 0.017 & 0.014 & 0.020 & 0.010 & 0.023 \\
fasttext-wikinews & 0.010 & 0.008 & 0.015 & 0.014 & 0.019 & 0.009 & 0.019 \\
bert-base & 0.007 & 0.003 & 0.006 & 0.008 & 0.009 & 0.006 & 0.003 \\
wordnet2vec & 0.018 & 0.012 & 0.027 & 0.016 & 0.023 & 0.017 & 0.040 \\
bert-large & 0.008 & 0.004 & 0.006 & 0.008 & 0.011 & 0.006 & 0.003 \\
elmo & 0.012 & 0.009 & 0.020 & 0.011 & 0.014 & 0.012 & 0.021\\\hline
\end{tabular}
\caption{Absolute mean squared error averaged over all features for each combination, i.e. averaged error of all eye-tracking features for each dataset.}
\end{table}

\begin{table}[h]
\centering
\begin{tabular}{lcccc}
\hline
\textbf{embeddings} & \textsc{Harry Potter} & \textsc{Nouncs} & \textsc{Alice} & \textsc{Pereira} \\\hline\hline
glove-50 & 0.005 & 0.204 & 0.036 & 0.044 \\
glove-100 & 0.015 & 0.220 & 0.069 & 0.055 \\
glove-200 & 0.007 & 0.224 & 0.036 & 0.050 \\
glove-300 & 0.008 & 0.224 & 0.038 & 0.050 \\
word2vec & 0.005 & 0.209 & 0.010 & 0.044 \\
fasttext-crawl & 0.002 & 0.194 & 0.009 & 0.039 \\
fasttext-wikinews & 0.001 & 0.185 & 0.004 & 0.037 \\
bert-base & 0.001 & 0.042 & 0.001 & 0.012 \\
wordnet2vec & 0.015 & 0.203 & 0.050 & 0.046 \\
bert-large & 0.001 & 0.055 & 0.001 & 0.013 \\
elmo & 0.001 & 0.135 & 0.001 & 0.034\\\hline
\end{tabular}
\caption{Absolute mean squared error averaged over all voxels in each fMRI dataset.}
\end{table}

\begin{table}[H]
\centering
\begin{tabular}{lcccc}
\hline
\textbf{embeddings} & \textsc{N400} & \textsc{Natural Speech} & \textsc{ZuCo} & \textsc{UCL} \\\hline\hline
glove-50 & 0.067 & 0.014 & 0.009 & 0.009 \\
glove-100 & 0.126 & 0.023 & 0.011 & 0.011 \\
glove-200 & 0.071 & 0.017 & 0.013 & 0.013 \\
glove-300 & 0.067 & 0.018 & 0.014 & 0.014 \\
word2vec & 0.047 & 0.017 & 0.012 & 0.012 \\
fasttext-crawl & 0.042 & 0.013 & 0.011 & 0.011 \\
fasttext-wikinews & 0.037 & 0.012 & 0.010 & 0.010 \\
bert-base & 0.014 & 0.005 & 0.006 & 0.006 \\
wordnet2vec & 0.089 & 0.020 & 0.015 & 0.015 \\
bert-large & 0.012 & 0.006 & 0.006 & 0.006 \\
elmo & 0.024 & 0.008 & 0.008 & 0.008\\\hline
\end{tabular}
\caption{Absolute mean squared error averaged over all electrodes in each EEG dataset.}
\end{table}

\newpage

\subsection{Correlations between datasets}

The following plots show example correlations between the prediction results within one modality, but across datasets. It shows correlations between different stimuli and different recording procedures.


\subsubsection*{Correlations between eye-tracking datasets}

\begin{figure}[!ht]
   \centering
   \subfloat[][]{\includegraphics[width=.45\textwidth]{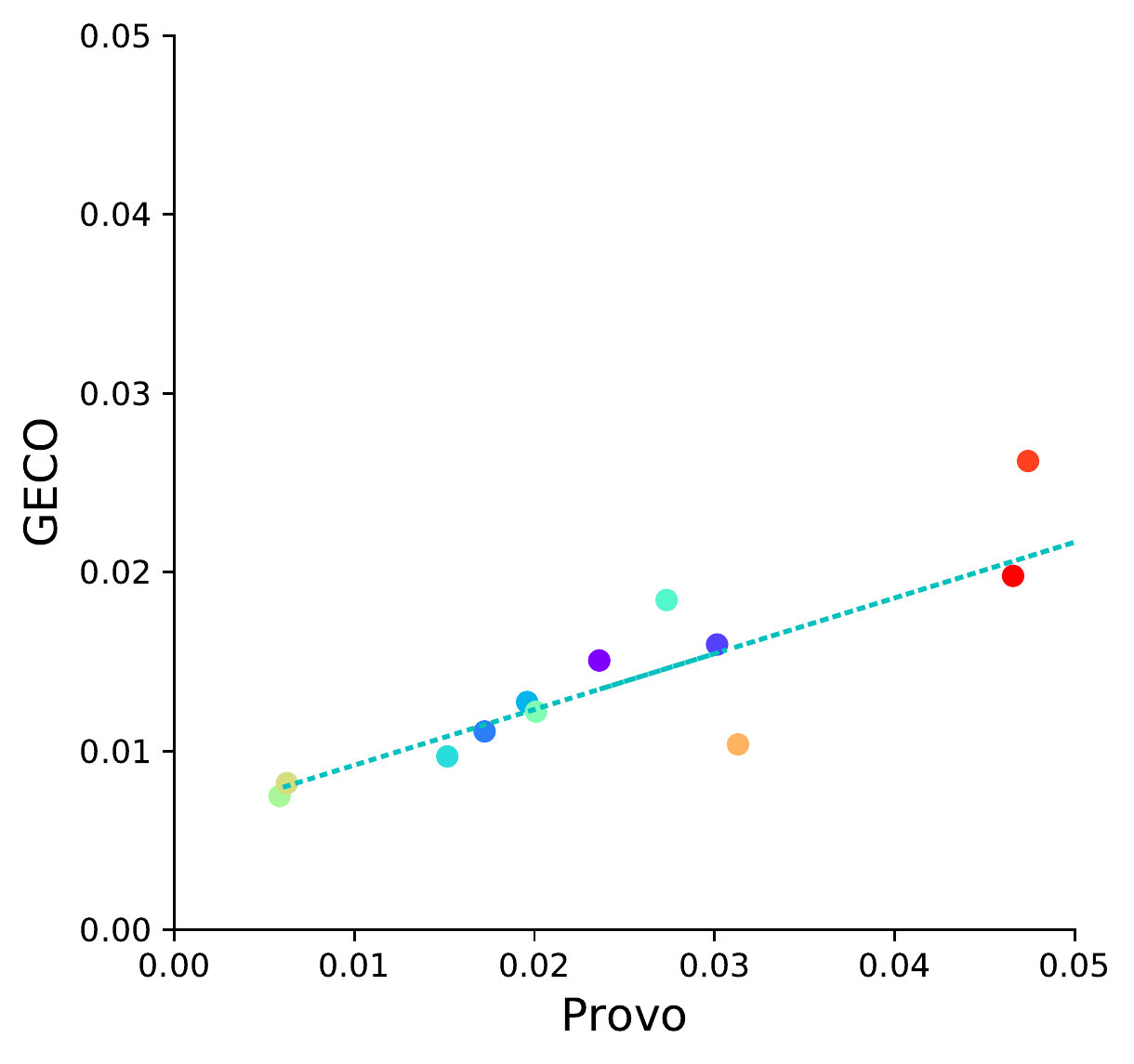}}\quad
   \subfloat[][]{\includegraphics[width=.45\textwidth]{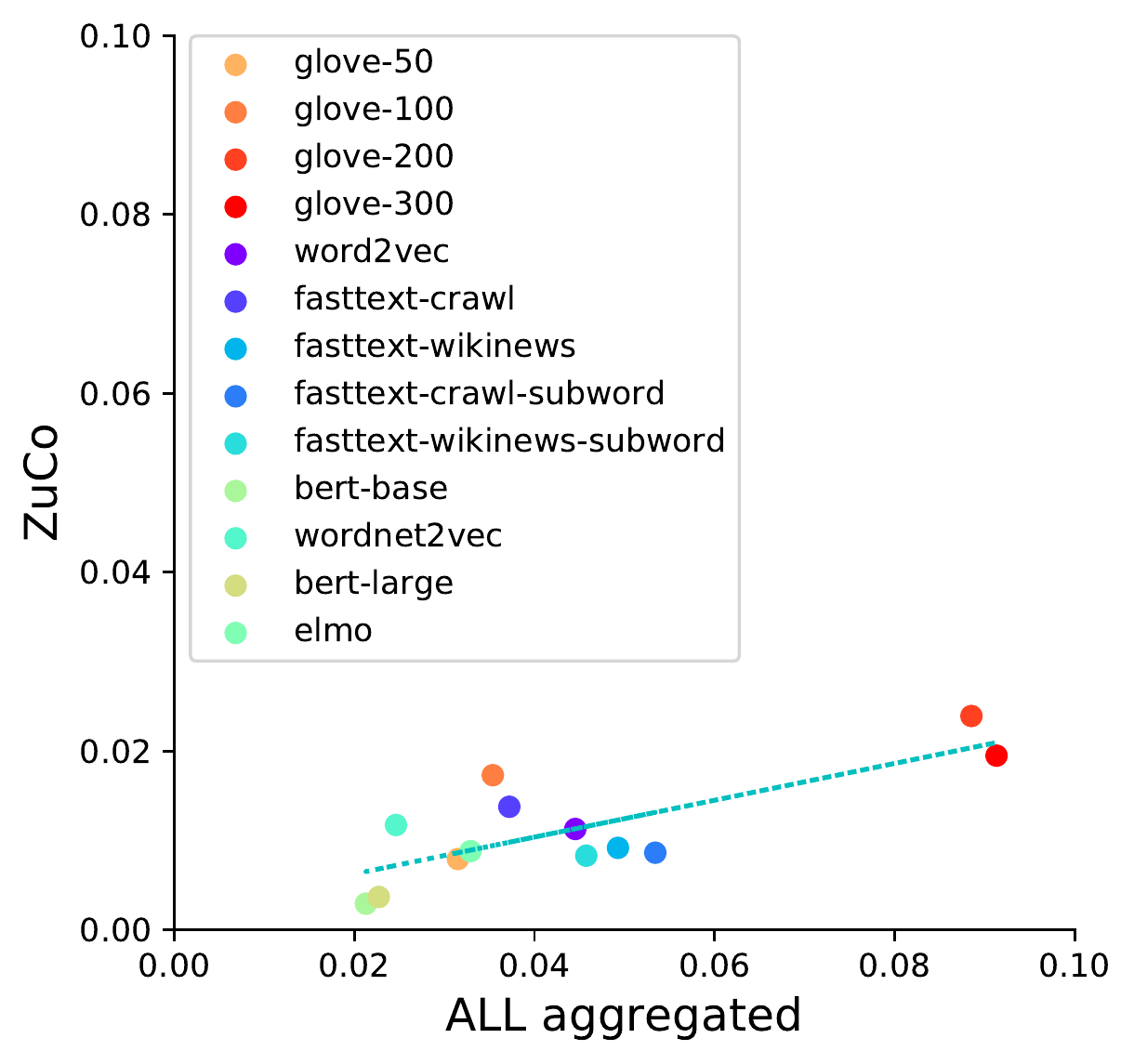}}\\
   \subfloat[][]{\includegraphics[width=.45\textwidth]{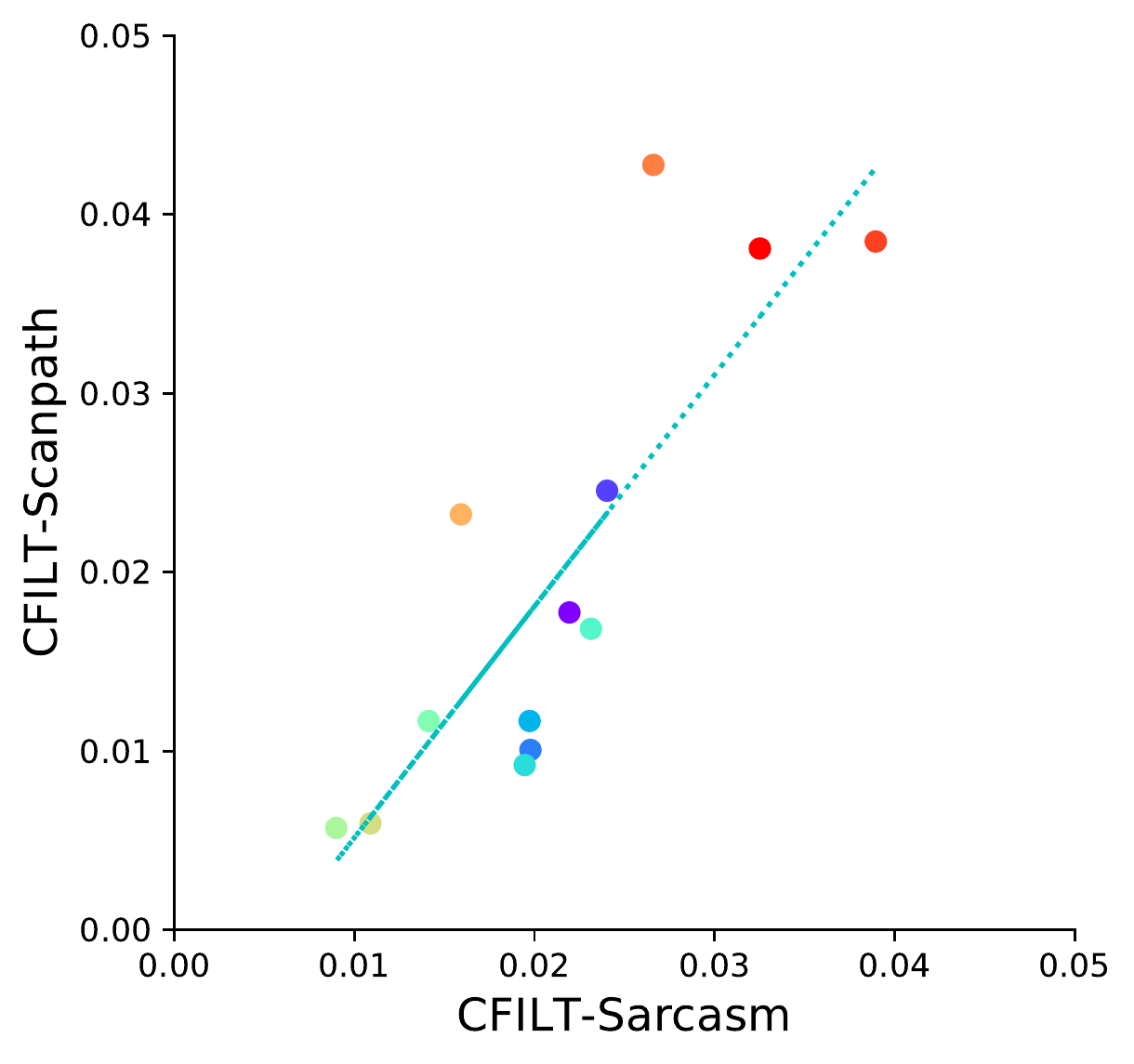}}\quad
   \subfloat[][]{\includegraphics[width=.45\textwidth]{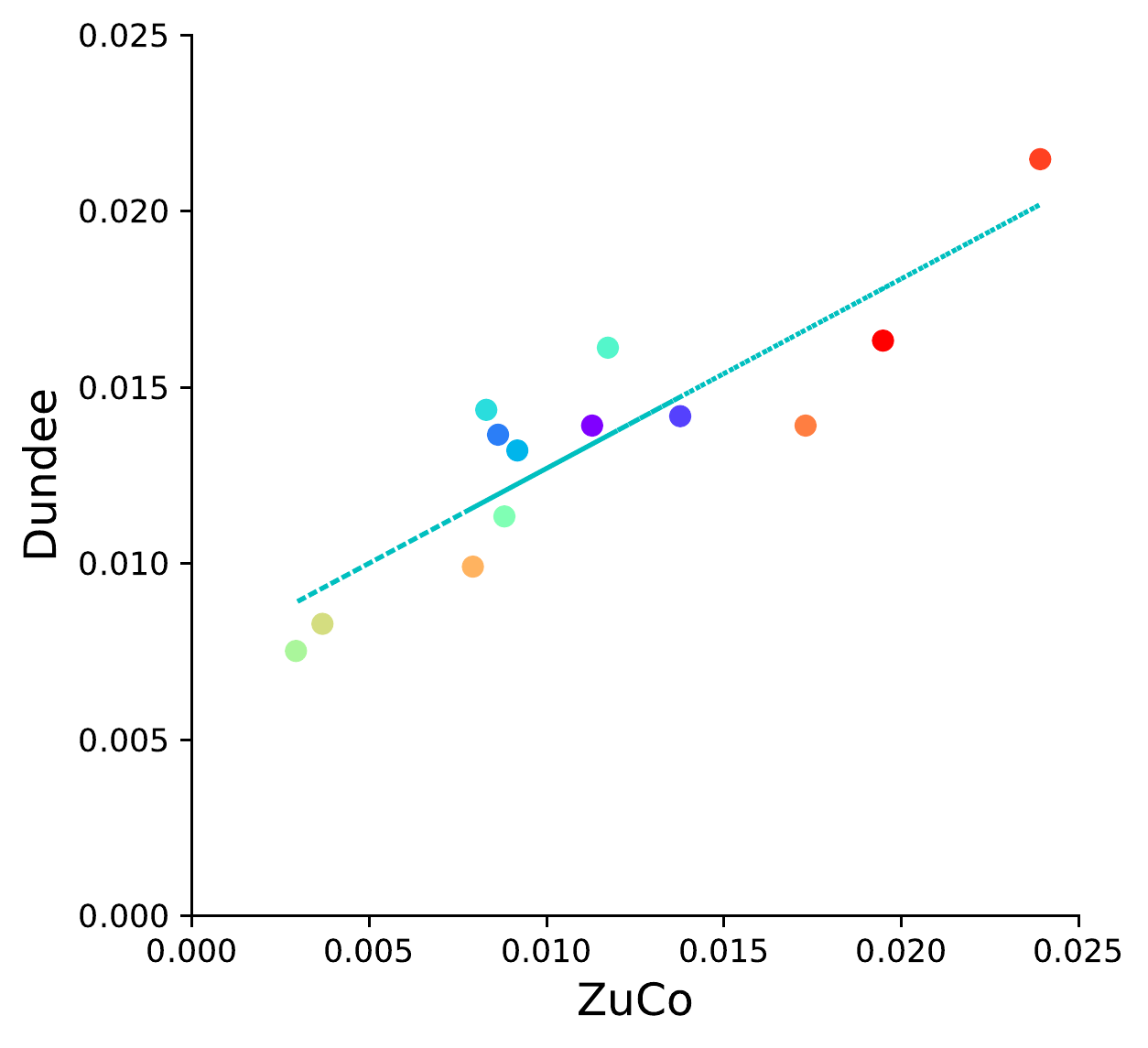}}
   \caption{Correlation plots between the prediction results of eye-tracking datasets.}
   \label{fig:correl-et}
\end{figure}

\newpage
\subsubsection*{Correlations between fMRI datasets}

\begin{figure}[H]
   \centering
   \subfloat[][]{\includegraphics[width=.45\textwidth]{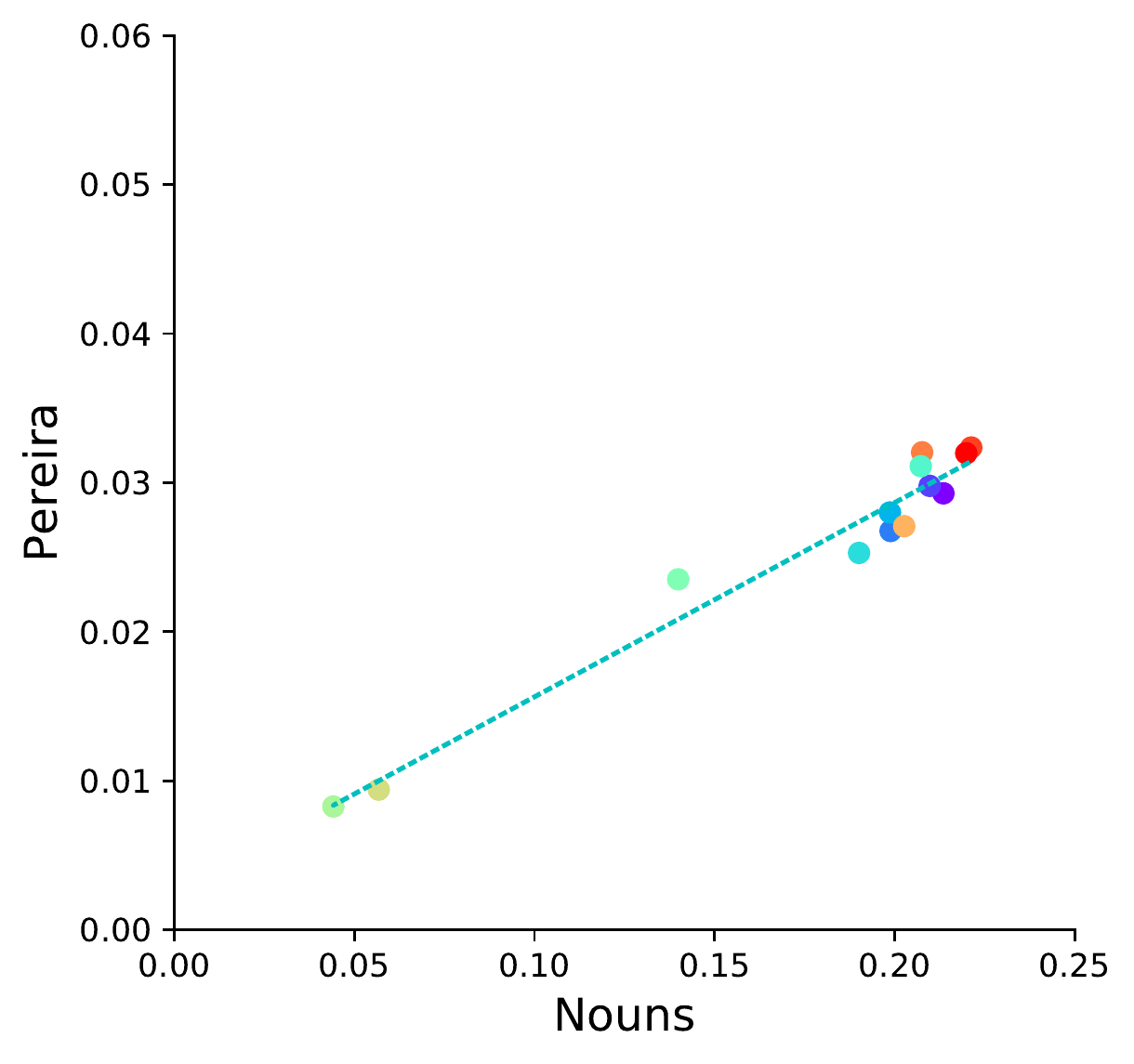}}\quad
   \subfloat[][]{\includegraphics[width=.45\textwidth]{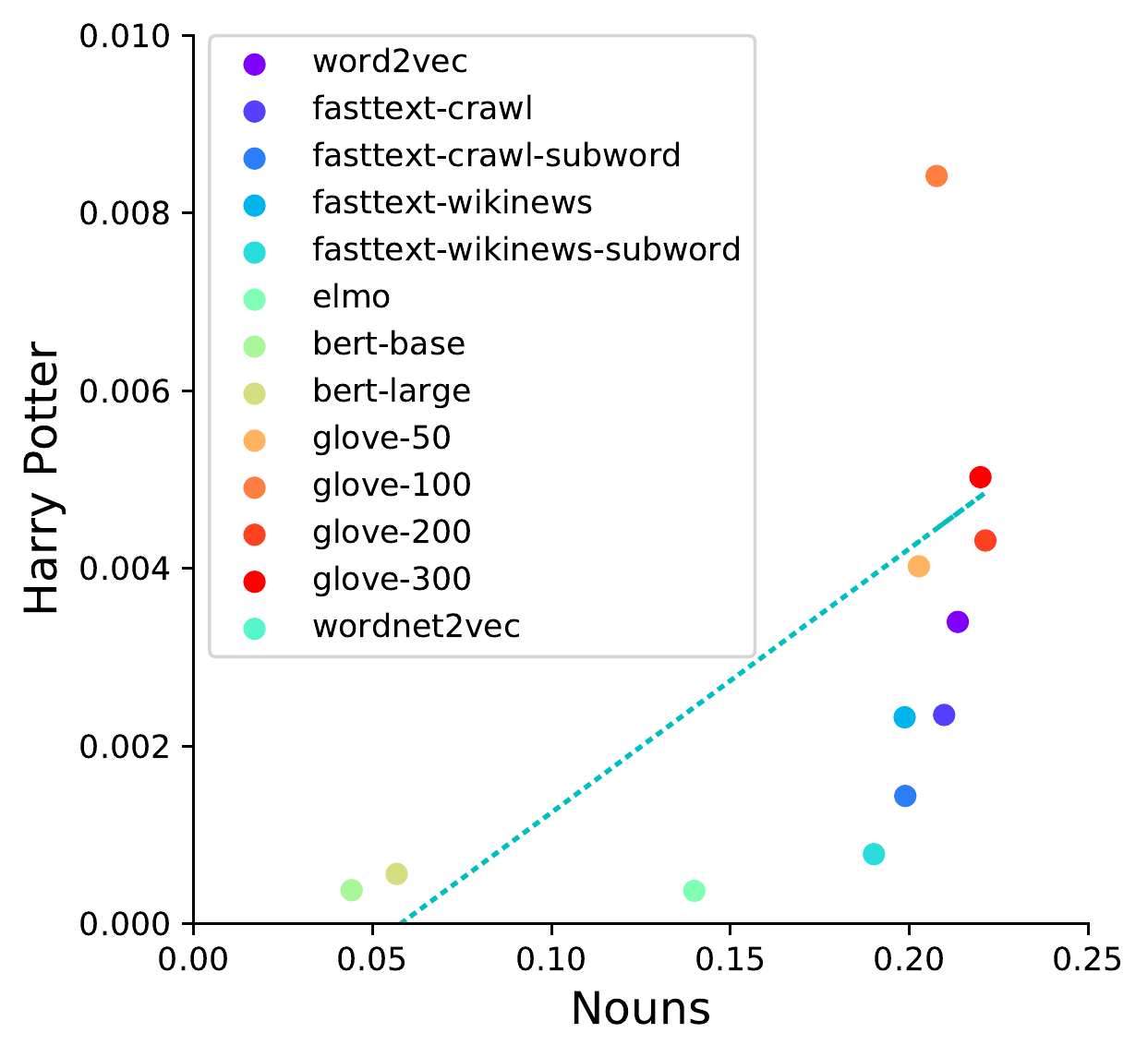}}\\
   \subfloat[][]{\includegraphics[width=.45\textwidth]{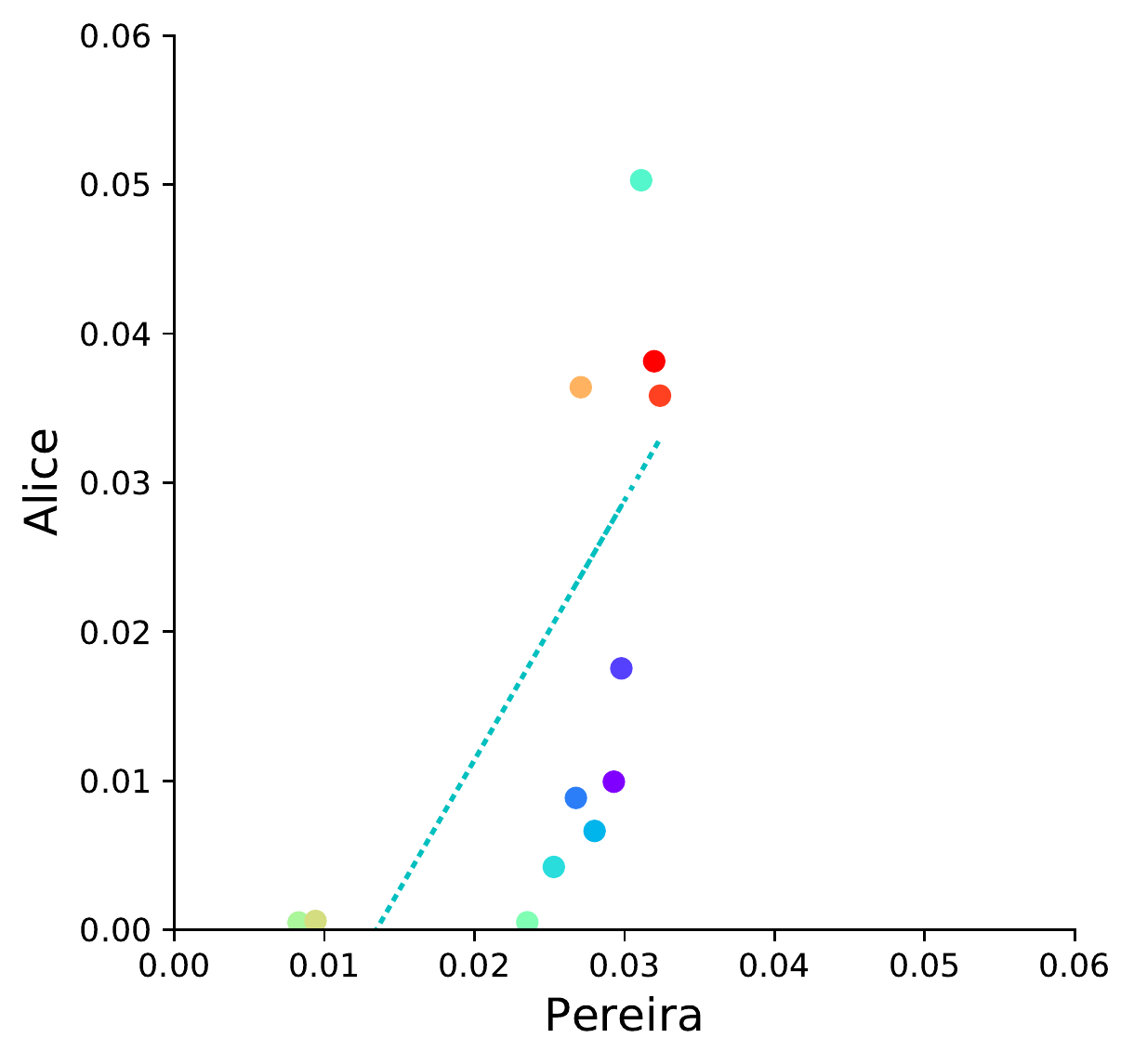}}\quad
   \subfloat[][]{\includegraphics[width=.45\textwidth]{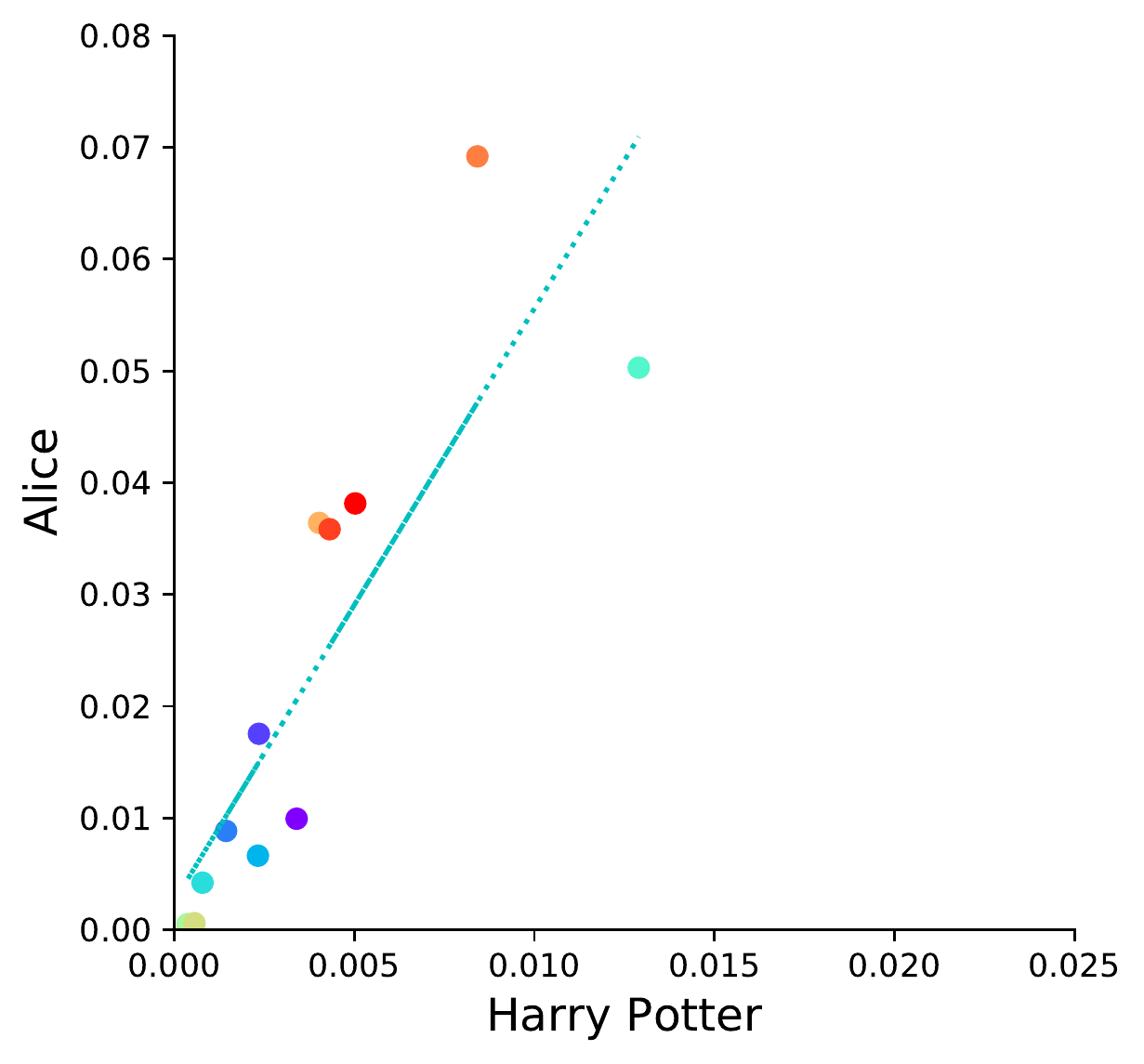}}
   \caption{Correlation plots between the prediction results of fMRI datasets.}
   \label{fig:correl-fmri}
\end{figure}

\end{document}